%% file: main.tex
\documentclass[10pt,twocolumn,letterpaper]{article}

\usepackage{cvpr}              
\input{preamble}
\definecolor{cvprblue}{rgb}{0.21,0.49,0.74}
\usepackage[pagebackref,breaklinks,colorlinks,allcolors=cvprblue]{hyperref}

\usepackage{utfsym}
\usepackage{fontawesome}
\usepackage{multirow}
\usepackage[most]{tcolorbox}
\tcbuselibrary{listings,breakable}
\definecolor{newgray}{gray}{0.4}
\usepackage{xspace}
\usepackage{setspace}
\usepackage{pifont}
\usepackage[table ]{ xcolor}

\def\paperID{*****} 
\def\confName{CVPR}
\def\confYear{2026}

\def\name{A4-Agent}

\title{\name : An Agentic Framework for Zero-Shot Affordance Reasoning}

\author{
Zixin Zhang\textsuperscript{1,4*} \quad
Kanghao Chen\textsuperscript{1,4*} \quad
Hanqing Wang\textsuperscript{1*} \quad
Hongfei Zhang\textsuperscript{1} \quad \\
Harold H. Chen\textsuperscript{1,4} \quad 
Chenfei Liao\textsuperscript{1,3} \quad
Litao Guo\textsuperscript{1} \quad
Ying-Cong Chen\textsuperscript{1,2\dag} \quad\\[5pt]
\normalsize$^{1}$HKUST(GZ)~~
\normalsize$^{2}$HKUST~~
\normalsize$^{3}$SJTU~~
\normalsize$^{4}$Knowin\\[5pt]
\normalsize{*Equal contribution \quad $^\dag$Corresponding author} \\
\normalsize{\textbf{\href{https://zixinzhang02.github.io/A4-Agent-page/}{[\faGlobe \ Project Page]} ~ \href{https://github.com/EnVision-Research/A4-Agent}{[\faGithub \ Github Repo]}}}
}

\begin{document}
\twocolumn[{
\maketitle
\begin{center}
    \noindent
    \includegraphics[width=\textwidth]{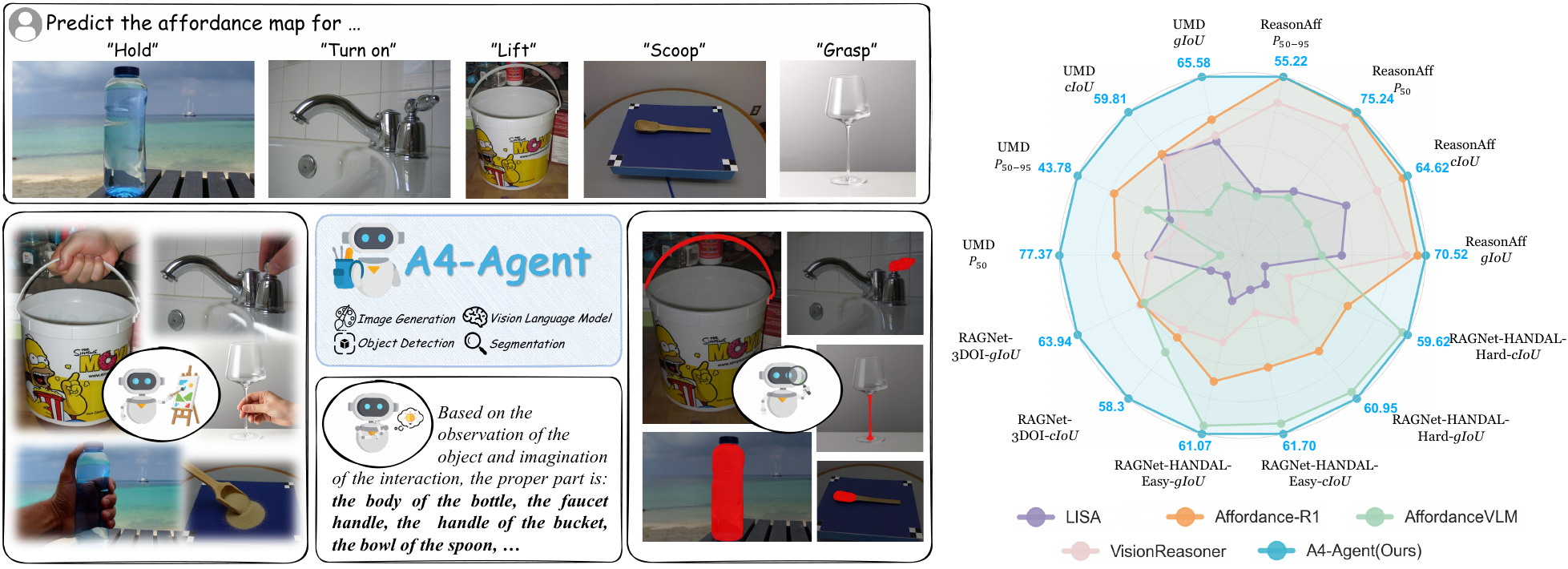}
    \captionof{figure}{\textbf{\textit{Left:}} Overview of \textbf{\name}~, an affordance-centric vision-language agent that predicts actionable regions based on complex task instruction. Given an observed object, A4-Agent integrates image generation, object detection, segmentation, and a vision-language model to imagine plausible interactions and localize the proper action-specific part. \textbf{\textit{Right:}} \name~ achieves state-of-the-art performance across multiple benchmarks with \textbf{\textit{zero-shot}} setting, surpassing baseline models that are specifically trained for affordance prediction task.}
    \label{fig:teaser}
\end{center}
\vspace{-0cm}
}]
\input{sec/0_abstract}

\input{sec/1_intro}
\input{sec/2_related}
\input{sec/3_method}
\input{sec/4_results}
\input{sec/5_conclusion}
\clearpage
{
  \small
  \bibliographystyle{ieeenat_fullname}
  \bibliography{main}
}
\clearpage
\input{sec/X_suppl}

\end{document}

%% file: sec/0_abstract.tex
\begin{abstract}
Affordance prediction, which identifies interaction regions on objects based on language instructions, is critical for embodied AI. Prevailing end-to-end models couple high-level reasoning and low-level grounding into a single monolithic pipeline and rely on training over annotated datasets, which leads to poor generalization on novel objects and unseen environments. In this paper, we move beyond this paradigm by proposing \name, a training-free agentic framework that decouples affordance prediction into a three-stage pipeline. Our framework coordinates specialized foundation models at test time: (1) a \textbf{Dreamer} that employs generative models to visualize \textit{how} an interaction would look; (2) a \textbf{Thinker} that utilizes large vision-language models to decide \textit{what} object part to interact with; and (3) a \textbf{Spotter} that orchestrates vision foundation models to precisely locate \textit{where} the interaction area is. By leveraging the complementary strengths of pre-trained models without any task-specific fine-tuning, our zero-shot framework significantly outperforms state-of-the-art supervised methods across multiple benchmarks and demonstrates robust generalization to real-world settings. 
\end{abstract}

%% file: sec/1_intro.tex
\section{Introduction}

Affordance, a concept describing the action possibilities that objects offer to agents, serves as a crucial bridge between visual perception and physical interaction. In the context of embodied AI and robotic manipulation, affordance prediction aims to identify specific regions of objects that enable task-relevant interactions based on natural language instructions. For instance, given the instruction ``open the refrigerator", a model must recognize the handle as the actionable region. This capability is fundamental to downstream applications including task planning~\cite{kumar2018visual}, robotic grasping~\cite{bahl2022human,hsu2023ditto}, and human-robot collaboration~\cite{chao2015hico,gkioxari2018detecting}, where understanding not just \textit{what} objects are present, but \textit{where} and \textit{how} to interact with them becomes essential for successful task execution.

Affordance prediction fundamentally requires two complementary capabilities: \ding{182} \textit{\textbf{high-level reasoning}}, interpreting natural language instructions and identifying task-relevant object parts, and \ding{183} \textit{\textbf{low-level grounding}}, precisely localizing these parts in pixel coordinates. 
Traditional approaches~\cite{li2023locate, luo2022learning, luo2023learning} mainly focused on grounding, treating it as a regression problem: given an affordance type, the model predicts an affordance map. However, such approaches lack high-level reasoning capabilities and therefore struggle to handle complex instructions. More recent studies~\cite{qian2024affordancellm, wu2025ragnet, affordance-r1} attempt to incorporate large language models (LLMs) and trained unified models that perform both reasoning and grounding. By fine-tuning on affordance datasets, these models are endowed with the ability to output affordance maps. However, such tightly coupled designs introduce several issues, including a trade-off between reasoning and grounding, limited generalization, and reduced flexibility, which ultimately hinder their applicability in real-world scenarios. This leads us to question:  \textit{despite the appeal of end-to-end systems, is entangling high-level reasoning and low-level grounding truly the right path forward for affordance prediction?}

In this paper, we present a preliminary exploration, \textbf{\name}, an agentic framework tailored to affordance prediction through training-free coordination of foundation models. Our key insight lies in decoupling the reasoning and grounding processes. We decompose the task into a three-stage pipeline, with each stage managed by a specialized expert leveraging powerful foundation models: \textbf{\textit{1) Dreamer}}: Drawing inspiration from human cognitive processes, the Dreamer initiates an imagination phase. It employs generative models to synthesize visual scenarios depicting \textit{how} an interaction would look (e.g., a hand grasping a handle, a door partially opening). \textbf{\textit{2) Thinker}}: The Thinker utilizes leading Vision-Language Models (VLMs) to interpret task instructions. Integrating visual observations with the imagined scenarios, it generates structured textual descriptions that specify \textit{what} to interact with. \textbf{\textit{3) Spotter}}: The Spotter orchestrates robust vision foundation models to execute precise spatial localization, pinpointing exactly \textit{where} the interaction area is within the visual input.

Remarkably, as shown in Fig.~\ref{fig:teaser}, by coordinating powerful pre-trained models without any task-specific training, our zero-shot framework \name~significantly outperforms current state-of-the-art supervised methods across multiple benchmarks and demonstrates robust generalization to real-world settings. To summarize, our main contributions are as follows:
\begin{itemize}
    \item We introduce A4-Agent, a training-free agentic framework that achieves superior performance and demonstrates strong zero-shot generalization capabilities.
    \item We validate a novel approach for affordance prediction by decoupling the reasoning and grounding processes. This allows for the integration of state-of-the-art models for each respective task, and we experimentally demonstrate the effectiveness of this method.
    \item We propose an Imagination-assisted affordance reasoning paradigm, showcasing the critical role of explicit imagination in the affordance reasoning process.
\end{itemize}

%% file: sec/2_related.tex
\section{Related Work}

\begin{figure*}[t]
    \centering
    \includegraphics[width=1\linewidth]{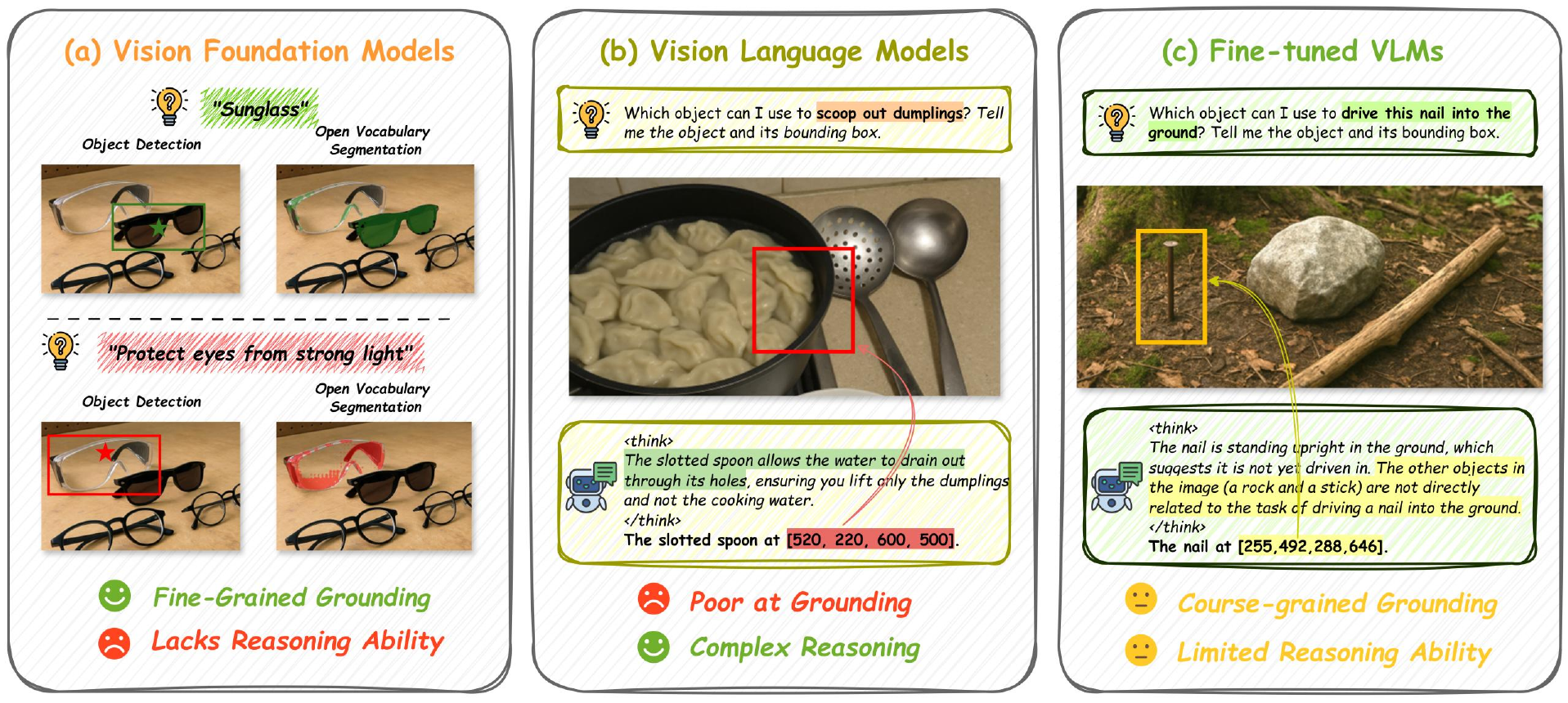}
    \vspace{-6mm}
    \caption{Vision Foundation Models are good at fine-grained grounding, but are poor at reasoning. Vision Language Models are good at reasoning, but are poor at visual grounding. Some works finetuned VLMs for better grounding ability, but both abilities are underwhelming.}
    \label{fig:motivation}
\end{figure*}
\paragraph{Affordance Learning.}
The concept of affordance, introduced by Gibson~\cite{gibson1977theory}, describes how agents perceive and interact with objects in their environment based on action possibilities. This foundational concept has inspired extensive research in affordance learning for robotic systems. Traditional approaches have explored various learning paradigms, including learning from human-object interaction (HOI) images~\cite{yang2023grounding,gao2024learning,shao2024great}, human demonstration videos~\cite{ma2025glover++}, and 3D perception through point clouds~\cite{deng20213d,qian2024affordancellm,yu2025seqafford,chu20253d,nguyen2023open} or 3D Gaussian Splatting~\cite{3DAffordSplat}. 

Recent advances have leveraged multimodal large language models (MLLMs) to enhance affordance understanding. For example, AffordanceLLM~\cite{qian2024affordancellm} and Seqafford~\cite{yu2025seqafford} introduce special tokens into the vocabulary and map affordance regions to token embeddings for segmentation outputs. LISA~\cite{lai2024lisa} extends this paradigm by incorporating reasoning capabilities for language-driven segmentation tasks. More recently, Affordance-R1~\cite{affordance-r1} employs reinforcement learning to enhance affordance reasoning and bounding box and key point grounding in MLLMs through process rewards. 
However, most of these methods adopt an end-to-end training paradigm that jointly optimizes reasoning and grounding capabilities within a single model architecture. They often face inherent trade-offs between reasoning complexity and spatial precision and exhibit poor generalization to novel scenarios.
In contrast, our method proposes a training-free agentic framework that coordinates foundation models to achieve zero-shot affordance prediction through explicit reasoning and grounding.

\paragraph{Multimodal Reasoning in MLLMs.}
MLLMs~\cite{yang2025qwen3,gpt4,llava} have demonstrated remarkable capabilities in visual understanding, generation, and multimodal reasoning. Recent advances have significantly enhanced their reasoning abilities through inference-time scaling. OpenAI o1~\cite{openaio1} extends the Chain-of-Thought (CoT)~\cite{cot} reasoning process to achieve superior performance, while DeepSeek-R1~\cite{guo2025deepseekr1} leverages reinforcement learning with GRPO~\cite{shao2024deepseekmath} to further advance reasoning capabilities. Building on these successes, several works~\cite{Shen2025VLMR1AS,seg-zero,Huang2025VisionR1IR} have expanded these reasoning paradigms to vision tasks, demonstrating the potential of enhanced reasoning in multimodal contexts.

Beyond text-based reasoning, emerging paradigms have explored reasoning with visual representations. 
VoT~\cite{wu2024minds} introduces textual imagery representations for dynamic reasoning. Benefiting from powerful generative models~\cite{generalworldmodelsurvey,zhang2025dualcamctrl,wan2025wan,guo2025comfymind,chen2025hierarchical}, approaches~\cite{li2025imagine,chen2025tivibench,chernThinkingGeneratedImages2025,guo2025video} attempt to leverage explicit visual imagination to assist in reasoning. 
These approaches show that generating intermediate visuals enhances reasoning and interpretability, offering valuable insights for affordance reasoning which demands complex spatial and interaction understanding.
Unlike existing end-to-end methods for affordance prediction, our agentic framework decouples reasoning from grounding, allowing for the seamless integration of these multimodal reasoning techniques.

%% file: sec/3_method.tex
\begin{figure*}[t]
    \centering
    \includegraphics[width=1\linewidth]{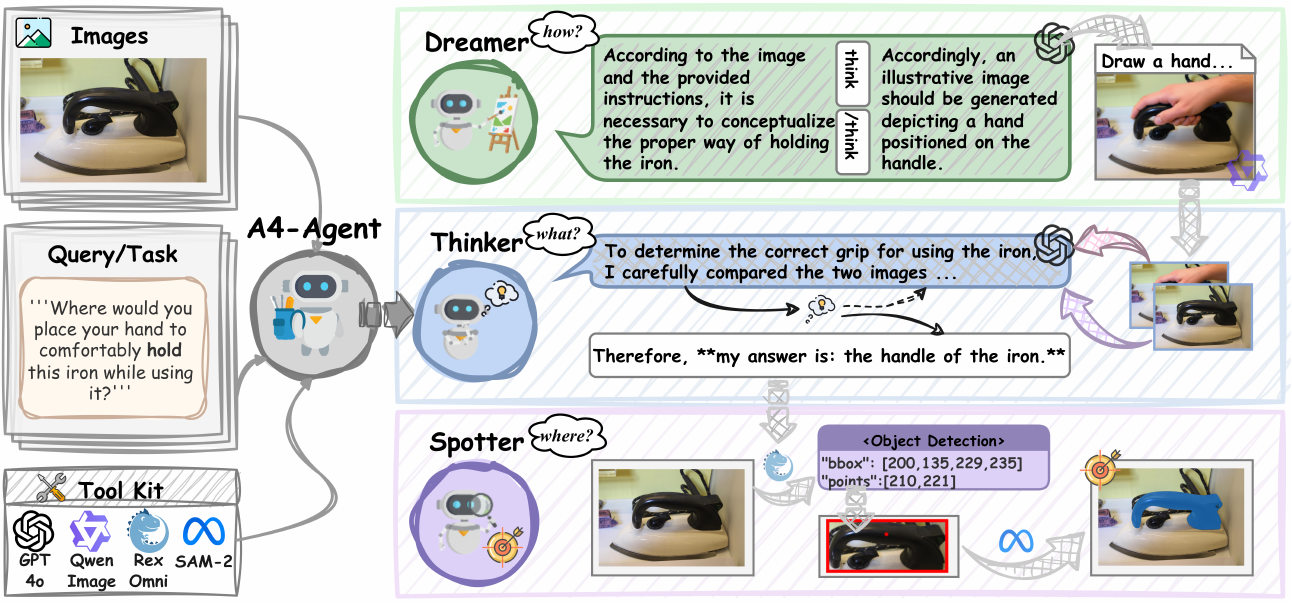}
    \vspace{-6mm}
    \caption{The pipeline of our \name framework, which decouples affordance prediction into three stages. \textbf{(1) Dreamer}: Imagines the interaction by generating a simulated image. \textbf{(2) Thinker}: Reasons over the original and simulated images to produce a textual description of the actionable object part. \textbf{(3) Spotter}: Takes this description to locate the part with bounding boxes and keypoints, then refines them into a precise segmentation mask.}
    \label{fig:pipeline}
\end{figure*}

\section{Motivation}
\label{sec:motivation}
Affordance prediction is a task that fundamentally requires two complementary capabilities: \textit{high-level reasoning} for interpreting instructions with object parts, and \textit{low-level grounding} for precisely localizing. As illustrated in Fig.~\ref{fig:motivation}~(a), while specialized vision foundation models excel at fine-grained localization, they lack the semantic understanding required to interpret complex task instructions. Conversely, as illustrated in Fig.~\ref{fig:motivation}~(b),  while recent MLLMs demonstrate impressive reasoning capabilities, they often produce coarse or inaccurate spatial predictions, rendering them insufficient for precise affordance prediction. 

Existing paradigm attempts to solve this dichotomy through monolithic end-to-end models. These approaches try to enhance reasoning models' grounding abilities~\cite{affordance-r1, visionreasoner, seg-zero} through training MLLMs on visual grounding data (\eg, bounding boxes, key points, masks). However, as Fig.~\ref{fig:motivation}~(c) shows, this tightly-coupled paradigm is less than ideal. They still introduces fundamental limitations: \ding{182} \textbf{Limited generalization:} training on limited datasets cannot cover the diversity of real-world scenarios, leading to brittleness on novel objects and environments; \ding{183} \textbf{Capability trade-offs:} optimizing for both reasoning and grounding simultaneously forces the model to balance different objectives, where improvements in one capability may degrade the other; \ding{184} \textbf{Poor flexibility:} the monolithic design prevents independent upgrades when more powerful foundation models emerge, requiring costly retraining of the entire system; and \ding{185} \textbf{Gap to closed-source models:} as these pipelines are restricted to open-source checkpoints, they cannot leverage the most capable closed-source models, thereby limiting the ceiling of reasoning ability.

Therefore, our work aims at exploring a fundamentally different approach: \textbf{\textit{decouple reasoning and grounding into specialized, coordinated agents}}. We argue that affordance prediction is inherently multi-stage. Rather than forcing a single model to master both capabilities, we design each component independently using state-of-the-art foundation models and orchestrate them through an agentic framework at test time. 

This paradigm shift can offer compelling advantages: \textbf{(I) Training-free generalization}, by leveraging pre-trained models' broad knowledge, the system generalizes to diverse scenarios without task-specific fine-tuning or expensive data collection; \textbf{(II) Modular specialization}, each component exploits the complementary strengths of different models and can be independently upgraded as better models become available; and \textbf{(III) Interpretable reasoning}, explicit intermediate steps make the decision-making process transparent and debuggable, facilitating error diagnosis and system refinement.

\section{\name:Agentic Affordance Reasoning}

\subsection{Problem Definition}
\label{subsec:problem_def}

We formulate affordance prediction as a visual grounding problem conditioned on natural language instructions. Given an input image $\mathbf{I} \in \mathbb{R}^{H \times W \times 3}$ along with a task description $\mathbf{T}$ (\textit{e.g.}, ``open the refrigerator''), the objective is to identify the affordance region $\mathcal{A}_{\text{ff}}$ that enables the specified interaction:
\begin{equation}
\mathcal{A}_{\text{ff}} = \mathcal{F}(\mathbf{I}, \mathbf{T}),
\end{equation}
where $\mathcal{A}_{\text{ff}}$ denotes the spatial region(s) where task-relevant interactions occur. Depending on downstream applications, this region can be represented as bounding boxes $\{\mathbf{B}_i\}_{i=1}^{N}$, key points $\{\mathbf{P}_i\}_{i=1}^{N}$, or segmentation masks $\{\mathbf{M}_i\}_{i=1}^{N}$. Following recent work~\cite{affordance-r1, wu2025ragnet}, we adopt segmentation masks as the primary representation for their pixel-level precision.

\subsection{Framework Overview}
\label{subsec:overview}
Building on the motivation outlined in Sec.~\ref{sec:motivation}, we introduce \name, a training-free, agentic framework for zero-shot affordance prediction that implements the decoupling principle. Unlike end-to-end models that directly regress $(\mathbf{B}, \mathbf{M})$ from $(\mathbf{I}, \mathbf{T})$, \name~first infers which object part requires interaction (reasoning) and then determines its location (grounding):

\begin{equation}
\mathcal{A}_{\text{ff}} = \mathbf{Ground}(\mathbf{Reason}(\mathbf{I}, \mathbf{T})).
\end{equation}

Specifically, the reasoning process follows a two-step pipeline: a Dreamer, which imagines \textit{how} the operation can be (Sec.~\ref{subsec:imagination}), and a Thinker, which decides \textit{what} part to the operation (Sec.~\ref{subsec:reasoning}). The grounding process is then handled by the Spotter to locate \textit{where} to operate using a coarse-to-fine approach (Sec.~\ref{subsec:grounding}): it initially identifies broad regions via bounding boxes and key points, which are then refined by a segmentation model to produce pixel-accurate masks. Overall framework is shown in Fig.~\ref{fig:pipeline}. 

\begin{figure*}[!h]
    \centering
    \includegraphics[width=1\linewidth]{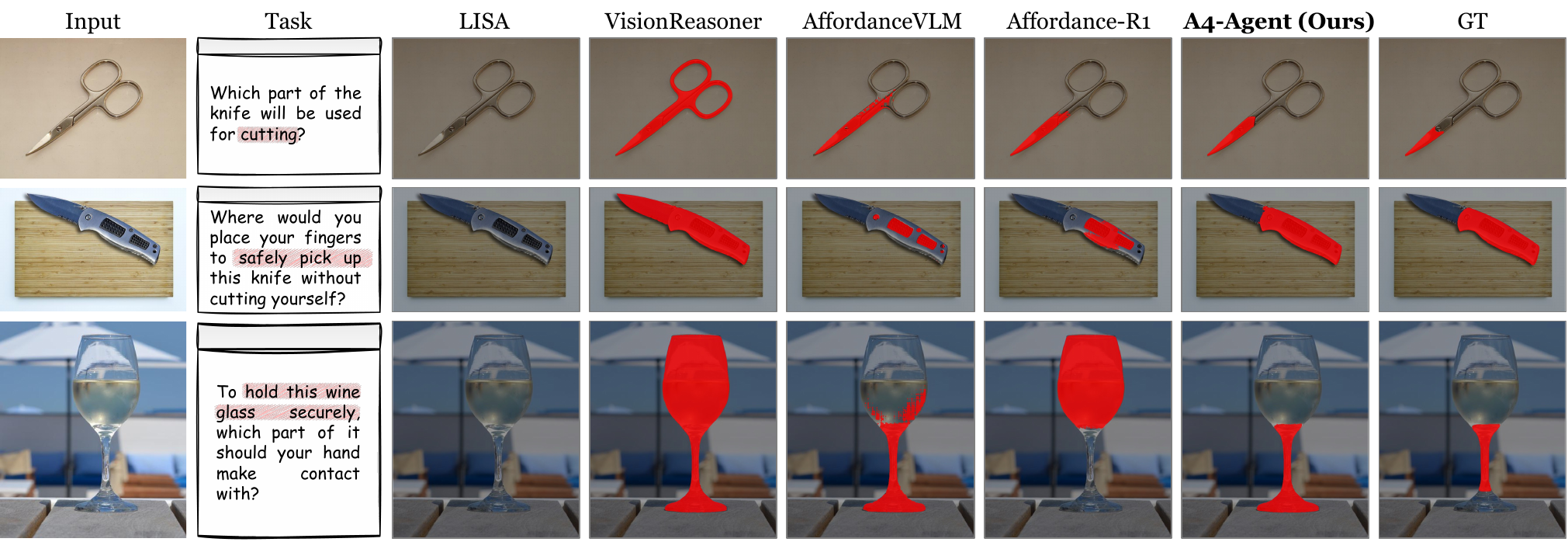}
    \vspace{-6mm}
    \caption{Qualitative comparison on ReasonAff dataset. Our method continuously predicts appropriate components according to task requirements, achieving results most consistent with ground truth and even surpassing Affordance-R1 specifically trained on this dataset.}
    \label{fig:reasonaff_vis}
\end{figure*}

\subsection{Dreamer: Imagine \textbf{\textit{how}} to Operate}
\label{subsec:imagination}
When humans reason about the affordances of a tool, they often begin by mentally simulating how the hand would interact with the tool and envisioning the broader usage scenario. Inspired by this process, we designed our Dreamer: rather than relying solely on text-based reasoning for affordance prediction, we first prompt the agent to use an image-generation module to visualize a plausible interaction state (\textit{e.g.}, a hand grasping a handle, a door being opened) based on the observation $\mathbf{I}$ and task $\mathbf{T}$. 

To construct the image editing prompt that drives this imagination step, we query a VLM with an instruction template applied to the pair $(\mathbf{I}, \mathbf{T})$. Formally,
\begin{equation}
\mathbf{T}_{\text{sim}} = \Phi_{\text{VLM}}(\mathbf{I}, \mathbf{T}; \tau),
\end{equation}
where $\Phi_{\text{VLM}}$ denotes the VLM and $\tau$ is our instruction template, which is detailed in the Appendix. This template asks the model to output a short, visually actionable description that: (\textbf{\textit{i}}) names the target object and the functional part visible in $\mathbf{I}$; (\textbf{\textit{ii}}) specifies the minimal interaction and contact configuration (\textit{e.g.}, ``a right hand grasping the vertical refrigerator handle''); and (\textbf{\textit{iii}}) avoids attributes not supported by the image. This yields concise prompts suitable for image editing and robust across varied scenes.
Following, we employ a generative model~\cite{qwen-image} to synthesize interaction scenarios. Given the original image $\mathbf{I}$ and a simulation prompt $\mathbf{T}_{\text{sim}}$ derived from the task instruction (\textit{e.g.}, ``a hand grasping the refrigerator handle''), the model produces an edited image $\mathbf{I}_{\text{sim}}$:
\begin{equation}
\mathbf{I}_{\text{sim}} = \mathcal{G}(\mathbf{I}, \mathbf{T}_{\text{sim}}),
\end{equation}
where $\mathcal{G}$ denotes the image generation model.
The imagined image $\mathbf{I}_{\text{sim}}$ explicitly highlights where interaction should occur by depicting plausible contact and motion cues and can further guides the agent in evaluating whether the action pattern is reasonable, thereby improving both the success rate and interpretability of affordance reasoning. This process can fully leverage the priors of the generative model, utilizing its understanding of interaction states to aid the reasoning model's reasoning process. Such seamless integration is made possible by our agentic framework.

\subsection{Thinker: Decide \textbf{\textit{what}} to Operate}
\label{subsec:reasoning}
The next step is to reason through the appropriate interactive areas in textual form. Given the original image $\mathbf{I}$, the imagined interaction image $\mathbf{I}_{\text{sim}}$, and the task $\mathbf{T}$, we prompt VLM with a preset template (see Appendix for the exact prompt) to perform three steps: (1) perceive key components and candidate interaction points in $\mathbf{I}$; (2) consult $\mathbf{I}_{\text{sim}}$ to infer contact/motion cues consistent with the affordance; (3) ground the actionable part back in $\mathbf{I}$ and return a compact, machine-readable specification.

The VLM returns two sections—\textit{Thinking} (free-form rationale) and \textit{Output} (a machine-readable JSON). We ignore the \textit{Thinking} section and parse only the \textit{Output} JSON with three fields: \texttt{"task"}, \texttt{"object\_name"}, and \texttt{"object\_part"}. The \texttt{object\_part} is phrased as ``the [object part] of the [object name]'' (\textit{e.g.}, ``the blade of the shears''). This yields a concise textual affordance description $\mathbf{D}$ specifying \textit{what} to interact with, without any spatial coordinates. This design reduces variance via explicit instruction-following, keeps the reasoning trace interpretable, and preserves modularity—stronger VLMs can be swapped in without retraining. 

\subsection{Spotter: Locate \textbf{\textit{where}} to Operate}
\label{subsec:grounding}
The Spotter translates semantic affordance descriptions from the reasoning process into precise pixel-level localizations. Given a textual description $\mathbf{D}$ (\textit{e.g.}, ``handle on the right refrigerator door''), we employ two complementary vision foundation models to achieve coarse-to-fine spatial grounding: an open-vocabulary detector for initial region identification, followed by a segmentation model for pixel-accurate mask refinement. This two-stage approach is motivated by the complementary strengths of existing foundation models: while segmentation models excel at producing precise boundaries, they require reliable visual prompts (\textit{e.g.}, boxes or points) rather than text, necessitating an initial detection step to bridge the semantic-geometric gap.

\begin{table}[t]
  \centering
    \caption{Quantitative results on ReasonAff. \textbf{\name~achieves SOTA performance in zero-shot manner without any training.}}
    \vspace{-2mm}
    \resizebox{\linewidth}{!}{
      \begin{tabular}{l|cccccc}
        \toprule
         Model & LLM &Reasoning &gIoU$\uparrow$  & cIoU$\uparrow$ & $P_{50}$$\uparrow$ & $P_{50-95}$$\uparrow$  \\
        \midrule
        VLPart~\cite{vlpart} & \usym{2613} & \usym{2613} & 4.21& 3.88& 1.31 & 0.85 \\
        OVSeg~\cite{ovseg} & \usym{2613} & \usym{2613} &16.52  & 10.59 & 9.89 & 4.12  \\
        SAN~\cite{SAN} & \usym{2613} & \usym{2613} &10.21 & 13.45 & 7.18 & 3.17  \\
        LISA-7B~\cite{lai2024lisa} & \checkmark & \usym{2613}&38.17 & 40.58 & 33.62 & 19.69 \\
        SAM4MLLM~\cite{chen2024sam4mllm} & \checkmark & \usym{2613} &45.51 & 33.64 & 43.48 & 22.79  \\
        AffordanceLLM~\cite{qian2024affordancellm} & \checkmark & \usym{2613} & 48.49  & 38.61 & 42.11 & 20.19 \\
        InternVL3-8B~\cite{internvl3} & \checkmark & \checkmark & 31.79 & 24.68 & 35.41 & 21.93  \\
        Qwen2.5VL-7B~\cite{bai2025qwen2} & \checkmark & \checkmark & 25.18 & 20.54 & 26.00 & 15.82  \\
        AffordanceVLM~\cite{wu2025ragnet} & \checkmark & \checkmark & 30.50  & 25.54 & 30.29 & 18.31 \\
        Seg-Zero~\cite{seg-zero} & \checkmark & \checkmark &59.26 & 48.03 & 61.33 & 45.87  \\
        Vision Reasoner~\cite{visionreasoner} & \checkmark & \checkmark &63.04 & 52.70 & 67.33 & 47.23  \\
        Affordance-R1~\cite{affordance-r1} & \checkmark & \checkmark &67.41& 62.72 & 74.50 & 55.22 \\
        \midrule
        \rowcolor{cyan!10}
        \textbf{\name~(Ours)} & \checkmark & \checkmark &\textbf{70.52}& \textbf{64.62} & \textbf{75.24} & \textbf{55.22} \\
        \bottomrule
      \end{tabular}
  }
\label{tab:main_result}
\end{table}

\paragraph{Open-Vocabulary Detection.}
We begin by using Rex-Omni~\cite{rex-omni}, a state-of-the-art open-vocabulary object detector, to perform initial spatial localization from textual descriptions. Given textual description $\mathbf{D}$ provided by Thinker, Rex-Omni outputs: \textbf{Bounding Boxes} $\{\mathbf{B}_i\}_{i=1}^{N}$: Rectangular regions that coarsely enclose the affordance parts. \textbf{Key Points} $\{\mathbf{P}_i\}_{i=1}^{N}$: Representative spatial anchors within each affordance region (\textit{e.g.}, the center of a handle).

\paragraph{Fine-Grained Segmentation with SAM.}
We then pass the bounding boxes ${\mathbf{B}_i}$ and key points ${\mathbf{P}_i}$ predicted by Rex-Omni as prompts to SAM, which generates detailed segmentation masks $\{\mathbf{M}_i\}_{i=1}^{N}$ that delineate the precise boundaries of the affordance regions. This prompt-based approach requires no additional training, directly leveraging SAM's powerful generalization capabilities developed through large-scale pretraining. The final affordance prediction aggregates multi-granular spatial information:
\begin{equation}
\mathcal{A}_{\text{ff}} = \{(\mathbf{B}_i, \mathbf{P}_i, \mathbf{M}_i)\}_{i=1}^{N},
\end{equation}
providing comprehensive spatial representations suitable for a variety of downstream applications—coarse bounding boxes for rapid scene understanding, key points for interaction targeting, and fine segmentation masks for precise manipulation planning.

In our Spotter, each model capitalizes on its strengths, and both can be independently upgraded as improved models emerge, without the need for end-to-end retraining.

%% file: sec/4_results.tex
\section{Experiment}

\subsection{Experimental Settings}

\begin{table}[t]
\centering
\caption{Quantitative results on RAGNet-3DOI and RAGNet-HANDAL. \textbf{\name~achieves SOTA performance in zero-shot manner without any training.}}
\vspace{-2mm}
\resizebox{\linewidth}{!}{
    \begin{tabular}{l|c|cc|cc|cc}
    \toprule
    \multirow{2}{*}{Model} & \multirow{2}{*}{Zero-shot} & \multicolumn{2}{c|}{3DOI} & \multicolumn{2}{c|}{HANDAL-easy} & \multicolumn{2}{c}{HANDAL-hard} \\
    \cmidrule(lr){3-4}\cmidrule(lr){5-6}\cmidrule(lr){7-8}
     &  & gIoU$\uparrow$ & cIoU$\uparrow$ & gIoU$\uparrow$ & cIoU$\uparrow$ & gIoU$\uparrow$ & cIoU$\uparrow$ \\
    \midrule
    G-DINO~\cite{liu2023grounding}& \checkmark & 4.1   & 3.9   & 3.6   & 3.0   & 3.4   & 3.1   \\
    LISA~\cite{lai2024lisa}     & \checkmark & 12.3  & 8.1   & 15.5  & 11.9  & 12.3  & 8.1   \\
    GLaMM~\cite{glamm}            & \checkmark & 4.4   & 2.9   & 4.7   & 3.5   & 5.0   & 3.5   \\
    Vision-Reasoner~\cite{visionreasoner}  & \checkmark & 39.6 & 30.3 & 29.6 & 19.8 & 27.7 & 16.7 \\
    Affordance-R1~\cite{affordance-r1}    & \checkmark & 39.0 & 33.4 & 43.1 & 38.7 & 40.7 & 37.9 \\
    AffordanceVLM~\cite{wu2025ragnet}    & \usym{2613}   & 38.1  & 39.4  & 58.3  & 58.1  & 58.2  & 57.8  \\
    \midrule \rowcolor{cyan!10}
    \textbf{\name~(Ours)}    & \checkmark & \textbf{63.9} & \textbf{58.3} & \textbf{61.1} & \textbf{61.7} & \textbf{61.0} & \textbf{59.6} \\
    \bottomrule
    \end{tabular}}
    
    \label{tab:ragnet}
\end{table}
\paragraph{Implementation Details}
\name~is a training-free framework coordinating pre-trained foundation models. In our complete agent, the VLM we used is GPT-4o~\cite{hurst2024gpt}, the geneative model we used is Qwen-Image-Editing~\cite{qwen-image}. For the open-vocabulary object detection, we use Rex-Omni~\cite{rex-omni}; and SAM2-Large~\cite{ravi2024sam} for the segmentation.

\paragraph{Datasets}
We evaluate \name~on three quantitative benchmarks and a set of open-world images to assess both reasoning-aware affordance prediction and generalization to diverse scenarios. \textbf{Crucially, our framework is completely zero-shot—it has never been trained or fine-tuned on any of these datasets.}

\noindent\textbf{1) ReasonAff}~\cite{affordance-r1}: A reasoning-oriented dataset built upon Instruct-Part~\cite{wan2024instructpart} with complex instructions requiring deep semantic understanding. We use the test split containing 600 image-task pairs.

\noindent\textbf{2) RAGNet}~\cite{wu2025ragnet}: A large-scale reasoning-based affordance segmentation dataset. We evaluate on two subsets: RAGNET-3DOI and RAGNET-HANDAL, containing 3,018 image-task pairs in total.

\noindent\textbf{3) UMD Part Affordance}~\cite{umd}: A standard affordance dataset covering 17 object categories with 7 affordance types. Following prior work~\cite{affordance-r1}, we sample one-tenth of the frames, yielding 1,922 test images.

\noindent\textbf{4) Open-World Images}: To evaluate generalization beyond standard benchmarks (which focus mainly on kitchen and household scenes), we collect diverse images from PhysToolBench~\cite{zhang2025phystoolbench} and web sources for qualitative evaluation. 

\subsection{Quantitative Results}
\begin{figure*}[t]
    \centering
    \includegraphics[width=1\linewidth]{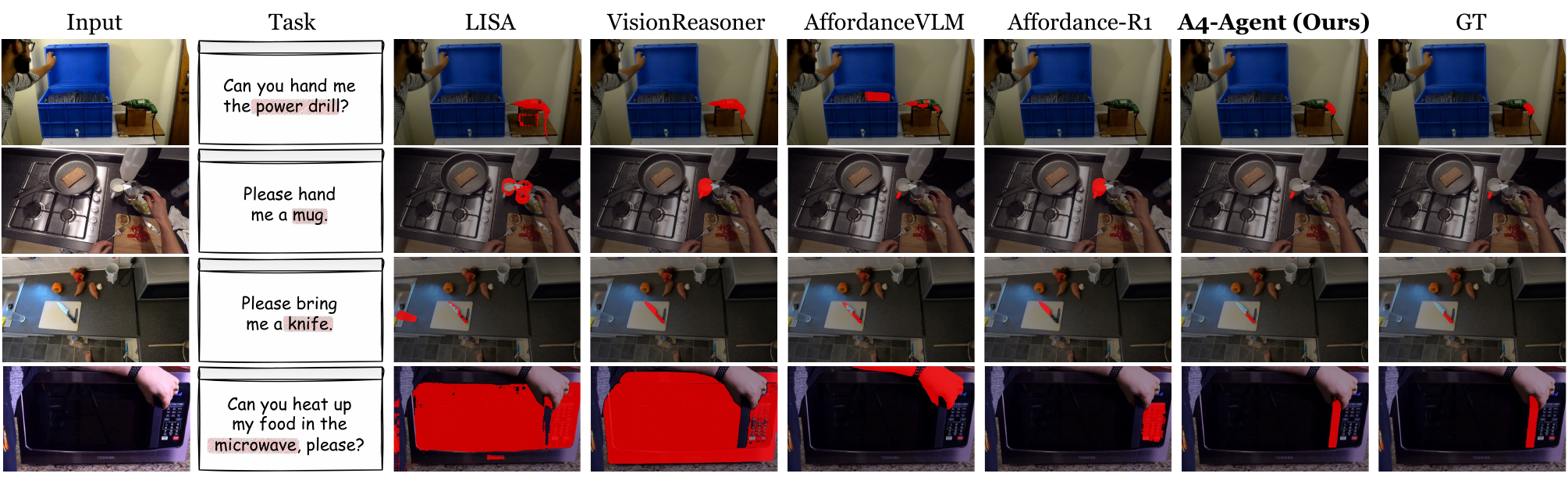}
    \vspace{-6mm}
    \caption{Qualitative comparison on RAGNet dataset. Our zero-shot method effectively reasons over task instructions to identify correct regions and precisely localize them with masks, closely matching ground truth. This outperforms baseline methods including AffordanceVLM trained on this dataset.}
    \label{fig:3doi_vis}
\end{figure*}

\paragraph{Results on ReasonAff Dataset.}
Table~\ref{tab:main_result} and Fig.~\ref{fig:reasonaff_vis} presents results on ReasonAff, which demands deep reasoning over implicit contextual instructions. \name~achieves state-of-the-art performance across all metrics without any training. Compared to supervised methods like AffordanceLLM (48.49 gIoU) and reasoning-enhanced approaches like Vision Reasoner (63.04 gIoU) and Affordance-R1 (67.41 gIoU), \name~reaches 71.83 gIoU, demonstrating superior reasoning ability and generalization.

This performance stems from three design principles. First, decoupling reasoning from grounding leverages complementary strengths—VLMs excel at semantic interpretation while specialized vision models provide precise localization. Second, the ``think-with-imagination'' mechanism grounds abstract instructions in synthesized visual representations, enhancing affordance understanding in complex scenarios. Third, unlike end-to-end models constrained by training data, our zero-shot approach naturally generalizes to ReasonAff's diverse instructions.

\noindent\textbf{Results on RAGNet Dataset.}~
Table~\ref{tab:ragnet} shows results on RAGNet, which focuses on reasoning-based affordance segmentation. Our framework demonstrates exceptional zero-shot performance, significantly outperforming all baselines. On 3DOI, \name~achieves 63.9 gIoU, surpassing Vision-Reasoner by over 24 points. This extends to HANDAL-hard and HANDAL-hard, where our \name~ also achieves highest score. Qualitative comparison is shown in Fig.~\ref{fig:3doi_vis}.

Critically, \name~even surpasses the supervised AffordanceVLM trained on this dataset, validating that \textbf{\textit{agentic coordination of foundation models outperforms task-specific fine-tuning}} for complex reasoning tasks. This stems from our ability to decompose abstract instructions into actionable steps and accurately ground them visually—a core benefit of our decoupled architecture.

\noindent\textbf{Results on UMD Dataset.}~
Besides the task of complex reasoning–intensive affordance prediction, \name~ also excels at the more traditional tasks of predicting affordances from action concepts. As shown in Table~\ref{tab:umd_results}, \name~ still achieves state-of-the-art performance, significantly outperforming the baselines by 15.53 gIoU, demonstrating a deep understanding of the many possible uses of different parts of common objects. This result is predictable, as this fundamental capability is arguably straightforward for powerful pre-trained models. This further supports our motivation: training-free coordination of specialized foundation models can exhibit strong generalization, as these models already possess sufficiently rich general knowledge.
\begin{figure*}[h]
    \centering
    \includegraphics[width=1\linewidth]{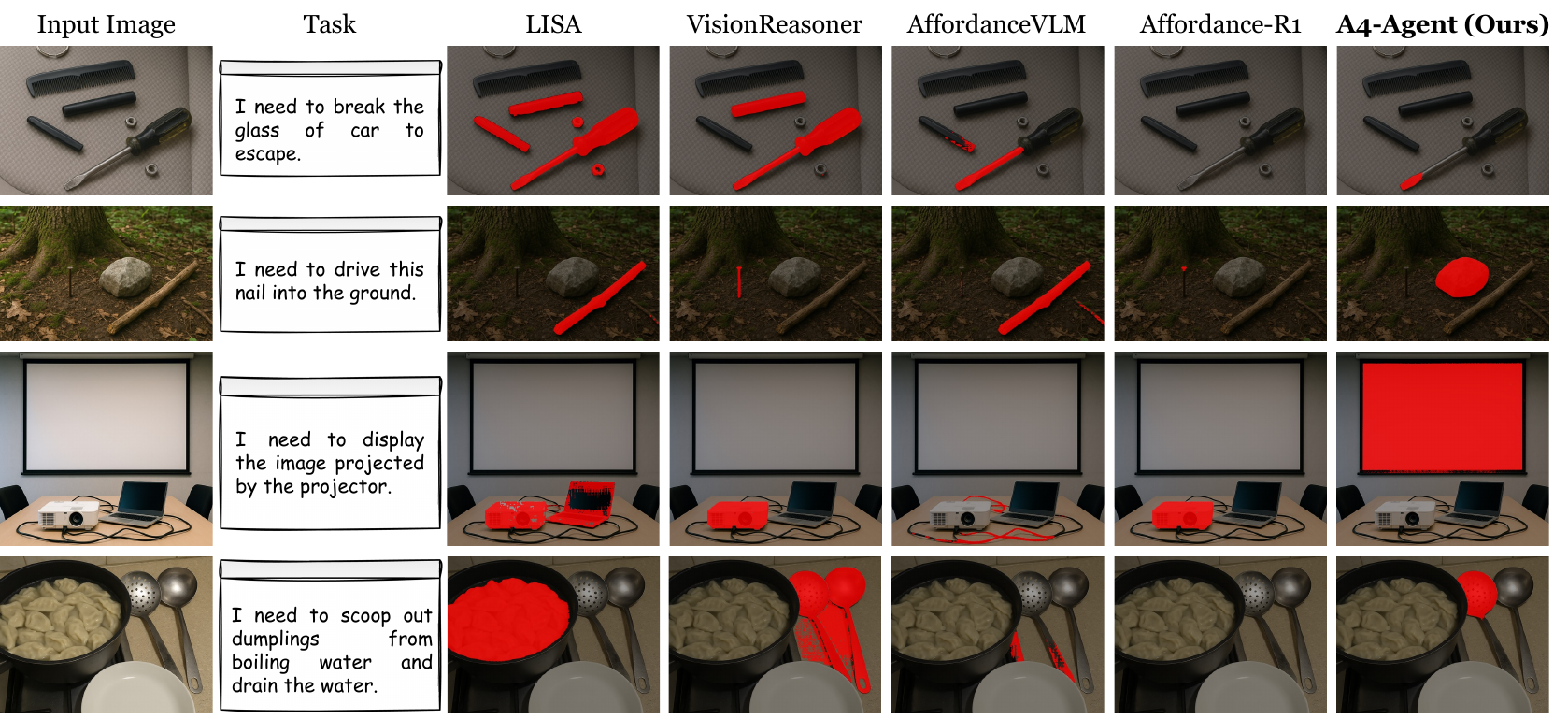}
    \caption{Qualitative results on open-world images. \name~demonstrates robust affordance reasoning across diverse scenarios, consistently produces reasonable regions based on complex instructions.}
    \label{fig:open_world}
\end{figure*}
\begin{table}[t]
  \centering
    \caption{Zero-shot results on UMD dataset. \name~outperforms fine-tuned methods without any training.}
    \resizebox{\linewidth}{!}{
      \begin{tabular}{p{3.0cm}|p{1.3cm} p{1.3cm} p{1.3cm} p{1.3cm}}
        \toprule
         Model  & gIoU$\uparrow$  & cIoU$\uparrow$ & $P_{50}$$\uparrow$ & $P_{50-95}$$\uparrow$  \\
        \midrule
        LISA-7B~\cite{lai2024lisa}  & 41.90 & 41.23 & 39.65 & 19.33 \\
        SAM4MLLM~\cite{chen2024sam4mllm}  & 12.40 & 8.41 & 4.12  & 0.05 \\
        AffordanceLLM~\cite{qian2024affordancellm}  & 43.11 & 38.97 & 41.56  & 22.36 \\
        Qwen2.5VL-7B~\cite{bai2025qwen2}  & 33.21 & 29.83 & 25.17  & 10.45 \\
        InternVL3-7B~\cite{internvl3} & 30.46 & 28.73 & 18.67  & 9.94 \\
        AffordanceVLM~\cite{wu2025ragnet} & 25.41 & 17.96 & 9.37  & 25.10 \\
        Seg-Zero~\cite{seg-zero}  & 44.26 & 39.30 & 39.93    & 16.53 \\
        Vision Reasoner~\cite{visionreasoner}  & 44.00 & 39.71 & 39.04  & 16.10 \\
        Affordance-R1~\cite{affordance-r1}  & 49.85 & 42.24 & 53.35  & 34.08 \\
        \midrule \rowcolor{cyan!10}
        \textbf{\name~(Ours)}  & \textbf{65.38} & \textbf{59.81} & \textbf{77.31}  & \textbf{43.78} \\
        \bottomrule
\end{tabular}

}
  
\label{tab:umd_results}
\end{table}

\subsection{Qualitative Results on Open-World Images}

To further validate the performance of A4-Agent on open-world scenarios, we perform a qualitative experiments of open-world images.
Fig.~\ref{fig:open_world} shows \name~'s strong performance across challenging scenarios: (1) \textbf{Novel objects}: Successfully identifying actionable regions on objects absent from standard benchmarks (\eg, digital equipment); (2) \textbf{Complex scenes}: Accurately identifying the most suitable part of a tool in complex environment (e.g., the tip of a screwdriver); (3) \textbf{Deep reasoning}: Using strong reasoning abilities to logically deduce the appropriate tools (e.g., a slotted spoon can be used to drain water, a rock can serve as a substitute for a hammer to drive nails).

Unlike baselines, which often fail on out-of-distribution objects, \name~maintains consistent performance by leveraging broad knowledge from web-scale pre-trained models. This confirms that training-free coordination has great potential for real-world application.

\subsection{Ablation Study}

\paragraph{Importance of Imagination in Affordance Reasoning.}
Table~\ref{tab:ablation_imagine} evaluates visual imagination's contribution. The imagination mechanism provides consistent improvements across all metrics for all base models. Notably, open-source Qwen-2.5-VL (7B) with imagination even outperforms closed-source GPT-4o using text-only reasoning. This validates that grounding reasoning in synthesized visual representations enhances affordance understanding, especially when textual descriptions alone are insufficient. The visual imagination mechanism serves as a bridge, allowing the reasoning model to effectively tap into and leverage the vast prior knowledge about interaction encapsulated within the generative model.

\begin{table}[t]
  \centering
  \caption{Ablation on Imagination on RAGNet-3DOI Dataset. *Affordance-R1 was fine-tuned from Qwen-2.5-VL-7B. T-w-I refers to think-with-imagination, which is the \textit{dreamer}.}
  \vspace{-2mm}
  \resizebox{\linewidth}{!}{
    \begin{tabular}{l|c|ll}
      \toprule
      Method & Reasoning backbone & gIoU$\uparrow$ & cIoU$\uparrow$ \\
      \midrule
      AffordanceVLM & LISA & 38.10 & 39.40 \\
      Affordance-R1 & Qwen-2.5-VL-7B* & 39.04 & 33.39 \\
      \midrule
      Ours w/o T-w-I & Qwen-2.5-VL-7B & 58.48 & 49.26 \\
      Ours w T-w-I  & Qwen-2.5-VL-7B 
      & 63.02 \;(\textcolor{ForestGreen}{$\uparrow\,4.54$})
      & 49.87 \;(\textcolor{ForestGreen}{$\uparrow\,0.61$}) \\
      Ours w/o T-w-I & GPT-4o & 62.30 & 54.43 \\
      Ours w T-w-I  & GPT-4o 
  & 63.94 \;(\textcolor{ForestGreen}{$\uparrow\,1.64$})
  & 58.30 \;(\textcolor{ForestGreen}{$\uparrow\,3.87$}) \\

      \bottomrule
    \end{tabular}
  }
  
  \label{tab:ablation_imagine}
\end{table}

\paragraph{Robustness to Different Components.}
We analyze \name's robustness to different component choices.

\noindent\textbf{Reasoning Backbone.} Table~\ref{tab:ablation_module} shows that replacing Qwen-2.5-VL with the more powerful GPT-4o significantly improves performance. This demonstrates \name's flexibility to seamlessly incorporate stronger foundation models as they become available.

\noindent\textbf{Segmentation Backbone.} Replacing SAM2-Large with smaller variants (SAM2-Base-Plus/Tiny) causes slight performance drops, but the framework remains highly effective and significantly outperforms baselines. The performance drop is also smaller than baseline method Affordance-R1. This underscores the robustness of our approach even with weaker grounding components.

\begin{table}[t]
  \centering
  \caption{Ablation on different components on RAGNet-3DOI Dataset. *AffordanceVLM is finetuned from LISA. SAM2-L,B,T denotes SAM2-Large, Base-plus, Tiny.}
  \vspace{-2mm}
  \resizebox{\linewidth}{!}{
  \begin{tabular}{l|c|c|c|c}
    \toprule
    Method & Reasoning & Segmentation & gIoU$\uparrow$ & cIoU$\uparrow$ \\
    \midrule
    AffordanceVLM & LISA* & LISA* & 38.10  & 39.40  \\
    \midrule
    \multirow{2}{*}{Affordance-R1} & Qwen-2.5-VL-7B* & SAM2-L & 39.04 & 33.39 \\
     & Qwen-2.5-VL-7B* & SAM2-T & 36.13 \;(\textcolor{BrickRed}{$\downarrow2.91$}) & 30.76 \;(\textcolor{BrickRed}{$\downarrow2.63$}) \\
    \midrule
    \multirow{4}{*}{Ours}
      & GPT-4o & SAM2-L & 62.30 \;(\textcolor{ForestGreen}{$\uparrow\,4.54$}) & 54.43 \; (\textcolor{ForestGreen}{$\uparrow\,5.17$}) \\
      & Qwen-2.5-VL-7B & SAM2-L 
        & 58.48 & 49.26 \\
      & Qwen-2.5-VL-7B & SAM2-B 
        & 56.84 \;(\textcolor{BrickRed}{$\downarrow1.64$})
        & 48.87 \;(\textcolor{BrickRed}{$\downarrow0.39$}) \\
      & Qwen-2.5-VL-7B & SAM2-T 
        & 56.32 \;(\textcolor{BrickRed}{$\downarrow2.16$})
        & 47.18 \;(\textcolor{BrickRed}{$\downarrow2.08$}) \\
    \bottomrule
  \end{tabular}
  }
  
  \label{tab:ablation_module}
\end{table}

%% file: sec/5_conclusion.tex
\section{Conclusion}
In this paper, we present \name, a novel training-free framework for affordance prediction. Our key contribution is decoupling the task into high-level reasoning and low-level grounding, enabling the use of vision-language models for semantic interpretation and vision foundation models for localization. We also introduce an imagination mechanism in reasoning, where a generative model visualizes potential interactions to improve the process. Extensive experiments show that this zero-shot approach outperforms supervised methods on challenging benchmarks and generalizes well to open-world scenarios. The success of A4-Agent highlights the potential of agentic coordination of foundation models for complex affordance prediction.

%% file: sec/X_suppl.tex
\clearpage
\setcounter{page}{1}
\maketitlesupplementary

\section{More Implementation Detail}

\subsection{Details of Baseline Methods}

\begin{figure}[h]
    \centering
    \includegraphics[width=1\linewidth]{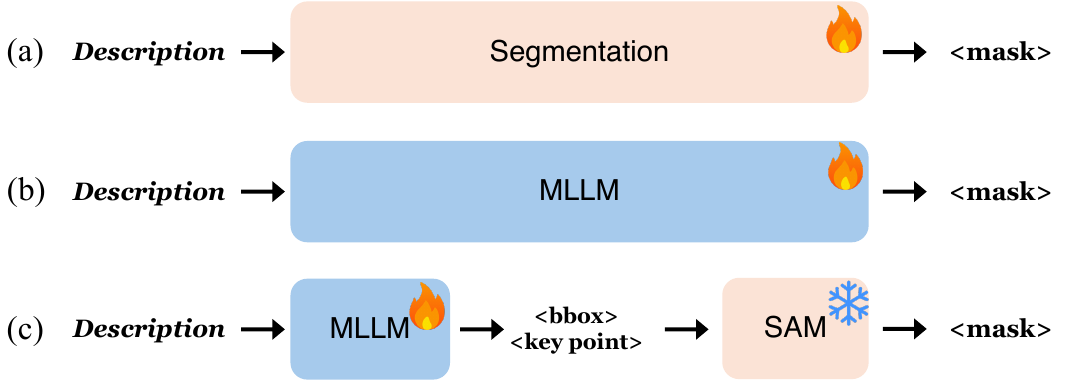}
    \caption{Illustration of different categories of baseline methods.}
    \label{fig:baselines}
\end{figure}

\noindent We comprehensively selected methods suitable for the affordance prediction task as our baselines, which fall into the following major categories:

\noindent\textbf{a) Open-Vocabulary Segmentation.}
These methods take open-vocabulary textual prompts as input and, given a description of an object, output a corresponding segmentation mask. Representative models include VLPart~\cite{vlpart}, OVSeg~\cite{ovseg}, SAN~\cite{SAN}, and Grounding-DINO(G-DINO)~\cite{liu2023grounding}.

\noindent\textbf{b) MLLM-Enhanced End-to-End Segmentation.}
These approaches fine-tune MLLMs to directly produce mask tokens and decode them into masks. Given a textual description, the models can generate segmentation masks in an end-to-end manner. This category includes AffordanceLLM~\cite{qian2024affordancellm}, AffordanceVLM~\cite{wu2025ragnet}, LISA~\cite{lai2024lisa}, SAM4MLLM~\cite{chen2024sam4mllm}, and GLaMM~\cite{glamm}.

\noindent\textbf{c) MLLM for Grounding + SAM for Segmentation.}
These methods follow a two-stage paradigm: the MLLM first performs task-aware grounding by predicting bounding boxes and keypoints for the target objects, and then a segmentation model (e.g., SAM2~\cite{ravi2024sam}) takes these as input to produce the final masks. Representative methods include Seg-Zero~\cite{seg-zero}, Vision Reasoner~\cite{visionreasoner}, and Affordance-R1~\cite{affordance-r1}, all of which are fine-tuned from open-source MLLMs. We also include open-source MLLMs with strong grounding ability such as Qwen-2.5-VL~\cite{bai2025qwen2} and InternVL-3~\cite{internvl3} for comparison.

\subsection{Evaluation Metrics}
Following the standard evaluation protocol in affordance prediction~\cite{affordance-r1,wan2024instructpart} and semantic segmentation~\cite{ravi2024sam,zhong2025omnisam}, we adopt four complementary metrics to comprehensively assess prediction quality including gIoU, cIoU, P@50, P@50:95. 

\noindent\textbf{gIoU (Generalized IoU)}: The average Intersection-over-Union across all images, measuring overall segmentation quality.

\noindent\textbf{cIoU (Cumulative IoU)}: The cumulative intersection over cumulative union, providing a dataset-level quality measure.

\noindent\textbf{P@50 (Precision at IoU=0.5)}: The percentage of predictions with IoU exceeding 0.5, evaluating high-quality predictions.

\noindent\textbf{P@50:95}: Average precision across IoU thresholds from 0.5 to 0.95 with 0.05 increments, providing a strict assessment of segmentation accuracy.

\subsection{System prompt of our Agent}

\begin{tcolorbox}[notitle, sharp corners, breakable, colframe=MidnightBlue!80, colback=gray!10, 
       boxrule=3pt, boxsep=0.5pt, enhanced, 
       shadow={3pt}{-3pt}{0pt}{opacity=1,newgray},
       title={Prompt for Dreamer}]\label{box:prompt1}
       \footnotesize
       \setstretch{1}
       {\fontfamily{pcr}\selectfont
\begin{lstlisting}
You are an "Imagination-driven Image-Editing Prompt Writer".
Input: (a) an image, (b) a TASK description.
Task: Based on the input image and TASK, imagine a person or another object interacting with a target object within the scene to do the task. Then, produce ONE concise, photorealistic image-editing prompt to be used by a downstream model to edit the image, depicting this interaction.

Requirements:
- The prompt must clearly describe the interaction, including the action, the state of the target object, and any necessary manipulators (e.g., a person's hand, a tool).
- Refer to the existing object and scene; do not replace them.
- Preserve the identity (shape, texture, color) of existing objects and the background. The camera viewpoint should remain unchanged.
- If introducing a person, describe the pose and action of the relevant body parts (e.g., a hand gripping a handle) realistically.
- Enforce physical plausibility: the scale, perspective, lighting, and shadows of any new elements must seamlessly match the original image.
- Ensure all occlusions are logical.

Output format:
- Output ONLY the editing prompt text (no JSON, no lists, no quotes, no explanations).
- Begin with ``Edit the input image to...'' and keep it to 1-3 sentences plus a short style clause (e.g., ``photorealistic, seamless inpainting'').
- End with "keep others unchanged".

The given TASK is:

\end{lstlisting}
}
\end{tcolorbox}

\begin{tcolorbox}[notitle, sharp corners, breakable, colframe=MidnightBlue!80, colback=gray!10, 
       boxrule=3pt, boxsep=0.5pt, enhanced, 
       shadow={3pt}{-3pt}{0pt}{opacity=1,newgray},
       title={Prompt for Thinker}]\label{box:prompt1}
       \footnotesize
       \setstretch{1}
       {\fontfamily{pcr}\selectfont
\begin{lstlisting}
Given the image of an object, the task is to decide which object to use and predict the part of the object that matches the provided task. The task instruction is "TASK". 
The first image is the original image of the object. The second image is the image of the object interacting with a person or another object in relation to the given affordance type for your reference.

**Follow these reasoning steps**:
1. Identify the key components of the object in the first image (e.g., shape, features, possible points of interaction).
2. Analyze the second image to understand how the object is interacting with a person or another object in relation to the given affordance type.
3. Go back to the first image and ground the part of the object in the image and output the result in a structured JSON format.

**Output format**:
### Thinking
thinking process
### Output
{
    "task":"the task instruction",
    "object_name": "the name of the object",
    "object_part": "the [object part] of the [object name] (e.g. the blade of the shears)"
}

\end{lstlisting}
}
\end{tcolorbox}

\section{More Exploratory Experiments}
Here we present additional exploratory experiments. For example, although Rex-Omni is used here as our object detection model, its backbone is an MLLM, which gives it much stronger language understanding capabilities than other traditional detection models. Motivated by this, we constructed, in an exploratory manner, a framework that uses only Rex-Omni and SAM. In terms of the A4-Agent, this corresponds to using only the Spotter module, without the preceding Dreamer and Thinker. We evaluate this variant on RAGNet-3DOI, and the results are shown below in the last line in Tab.~\ref{tab:more}.

\begin{table}[h]
    \centering
    \caption{More Exploratory results on RAGNet-3DOI.}
    \resizebox{\linewidth}{!}{
    \scriptsize
    \begin{tabular}{ccc|cc}
        \toprule
        Dreamer & Thinker & Spotter & gloU$\uparrow$ & cloU$\uparrow$ \\
        \midrule
        \checkmark & \checkmark & \checkmark & 63.94 & 58.30 \\
        \usym{2613}   & \checkmark & \checkmark & 62.30  & 54.43 \\
        \usym{2613}   & \usym{2613}   & \checkmark & 45.91 & 39.82 \\
        \bottomrule
    \end{tabular}}
    \label{tab:more}
\end{table}

Although Spotter itself has some reasoning ability, its backbone is only a small MLLM with limited reasoning capacity, which leads to suboptimal performance. This further validates our motivation: \textbf{\textit{by decoupling the reasoning and grounding processes, we can fully exploit their respective strengths and easily scale up the system to improve its overall performance.}}

\section{More Intermediate Results}
\begin{figure*}
    \centering
    \includegraphics[width=1\linewidth]{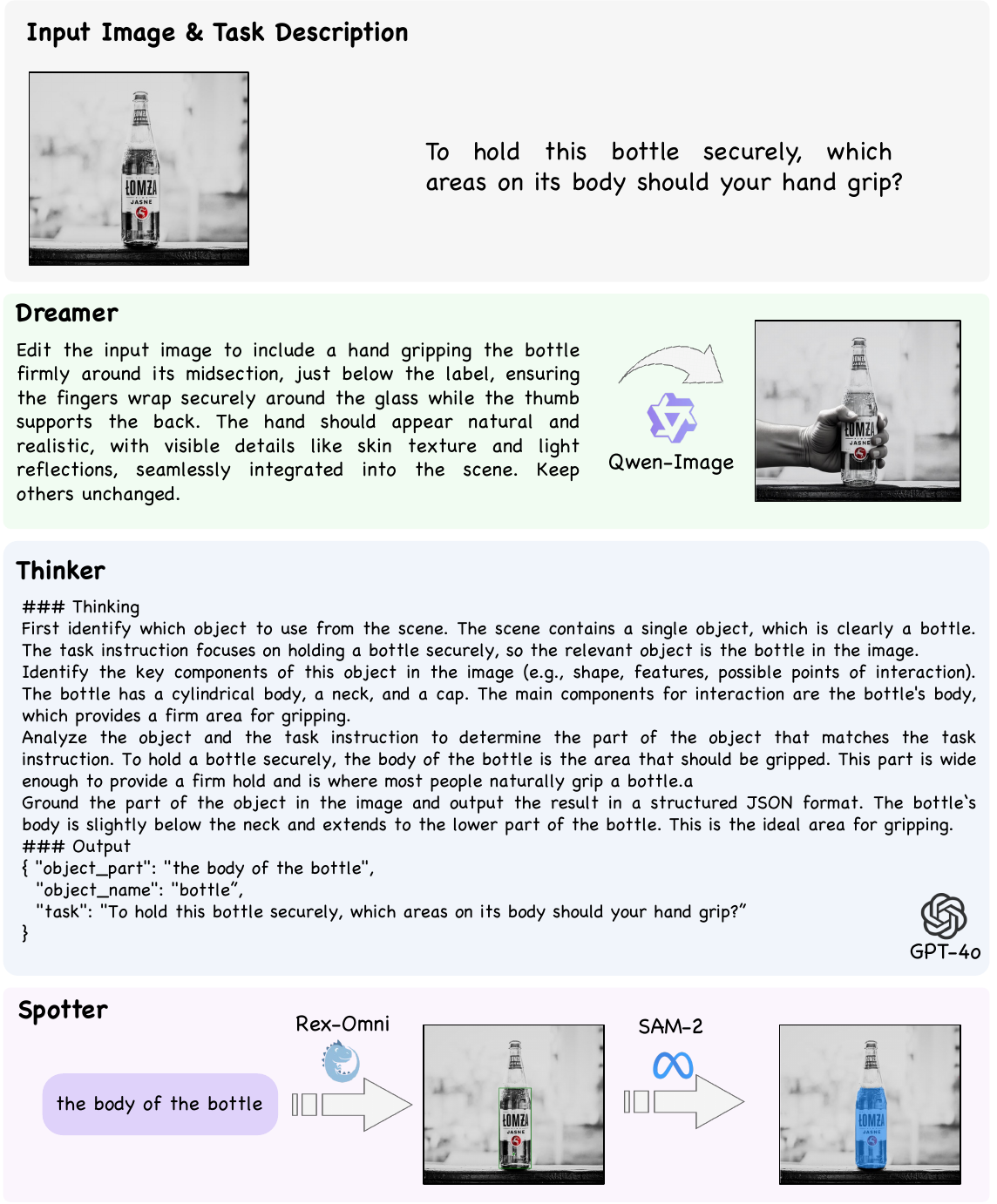}
    \caption{\textbf{Full Demonstration of Intermediate Results.} Sampled from results on the ReasonAff dataset.}
    \label{fig:case1}
\end{figure*}

\begin{figure*}
    \centering
    \includegraphics[width=1\linewidth]{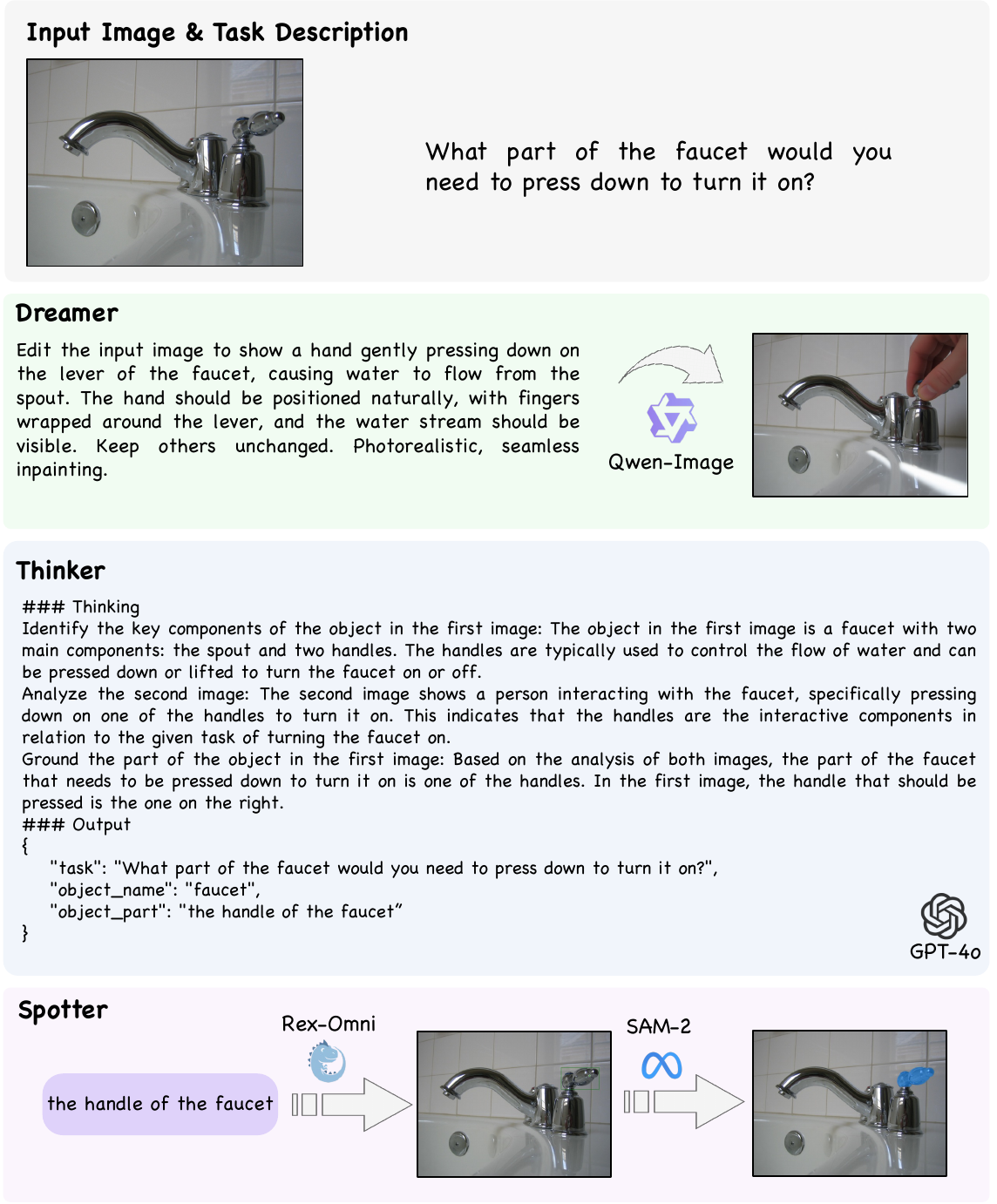}
    \caption{\textbf{Full Demonstration of Intermediate Results.} Sampled from results on the ReasonAff dataset.}
    \label{fig:case2}
\end{figure*}

\begin{figure*}
    \centering
    \includegraphics[width=1\linewidth]{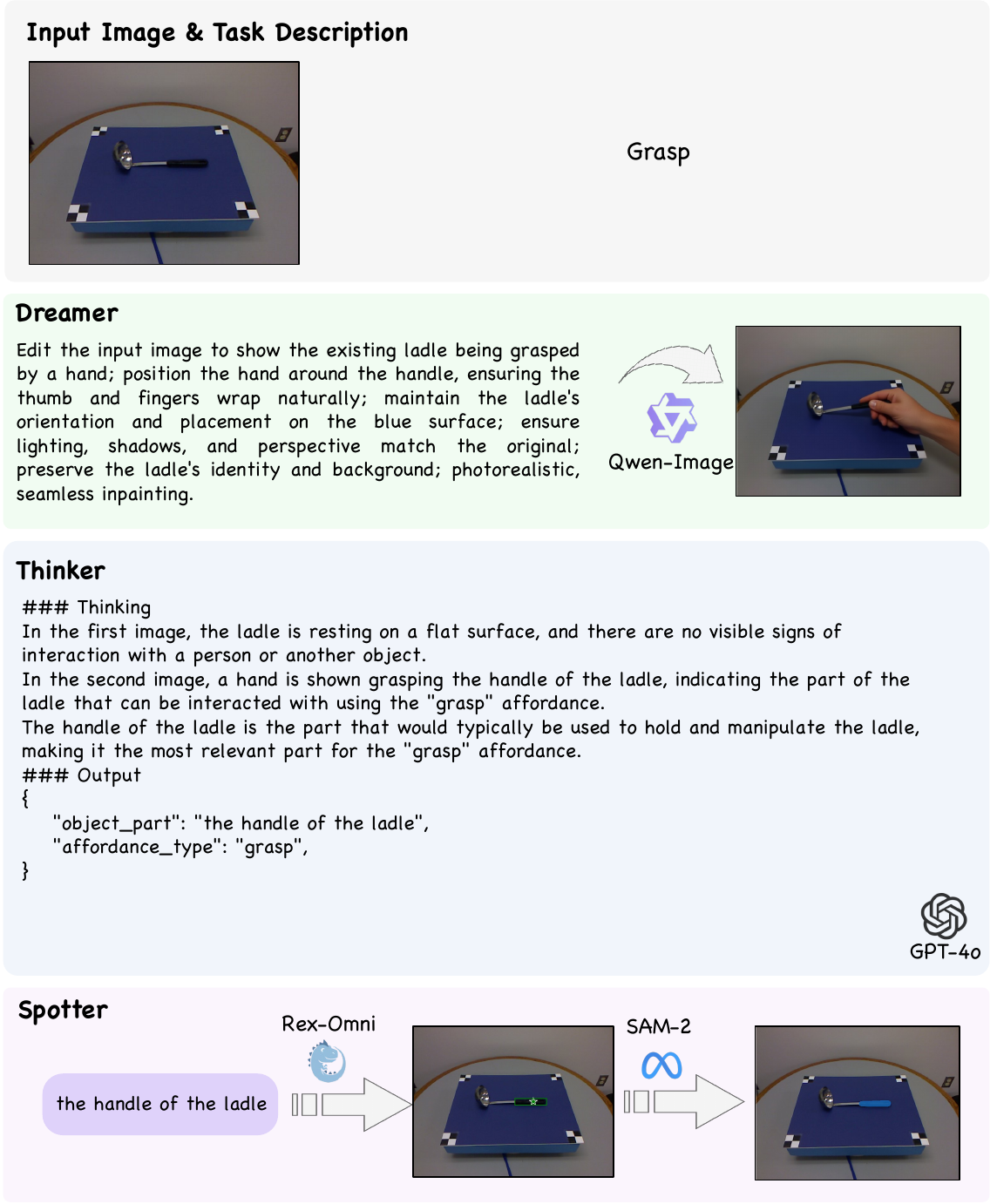}
    \caption{\textbf{Full Demonstration of Intermediate Results.} Sampled from results on the UMD dataset.}
    \label{fig:case3}
\end{figure*}

\begin{figure*}
    \centering
    \includegraphics[width=1\linewidth]{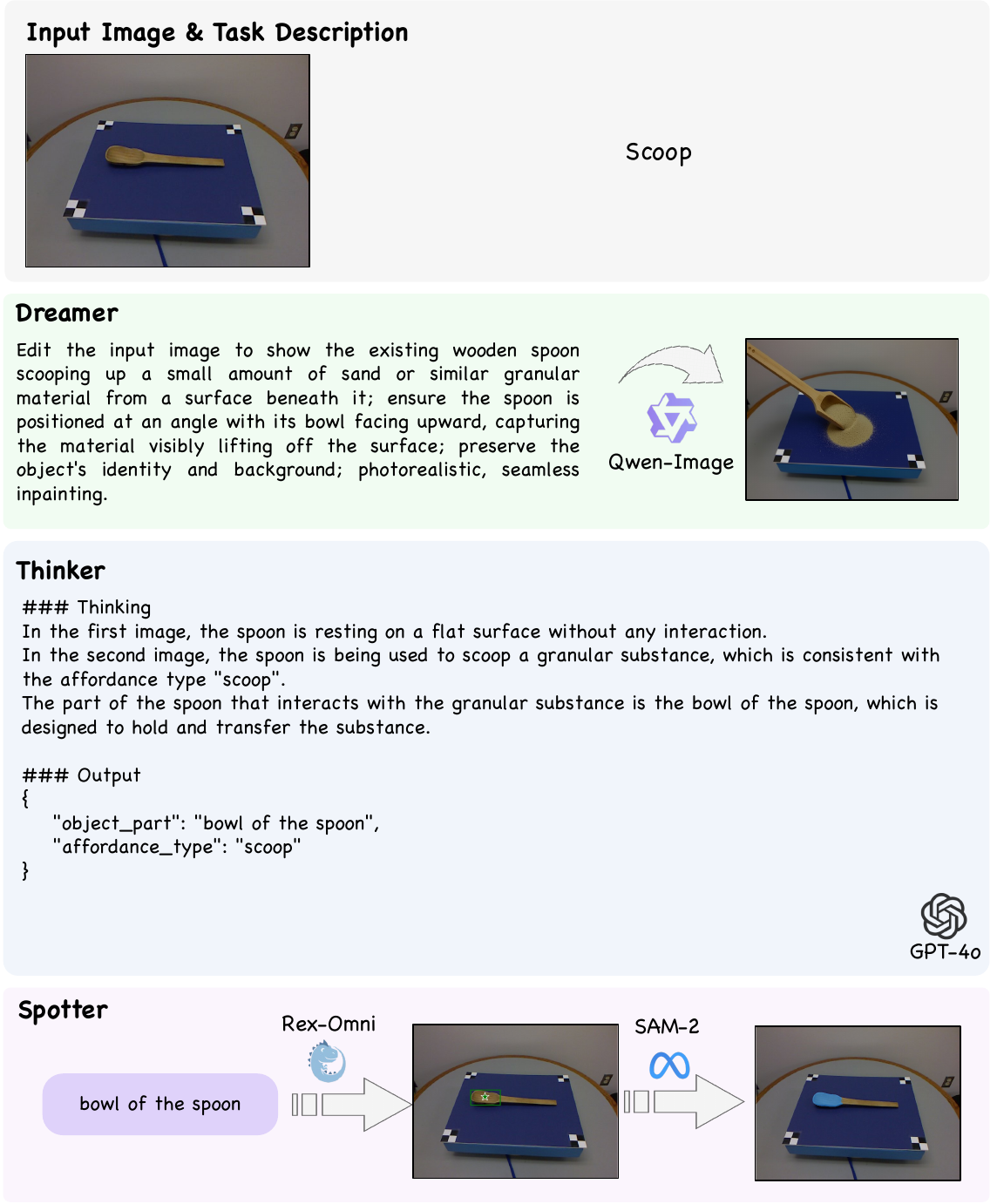}
    \caption{\textbf{Full Demonstration of Intermediate Results.} Sampled from results on the UMD dataset.}
    \label{fig:case4}
\end{figure*}

\begin{figure*}
    \centering
    \includegraphics[width=1\linewidth]{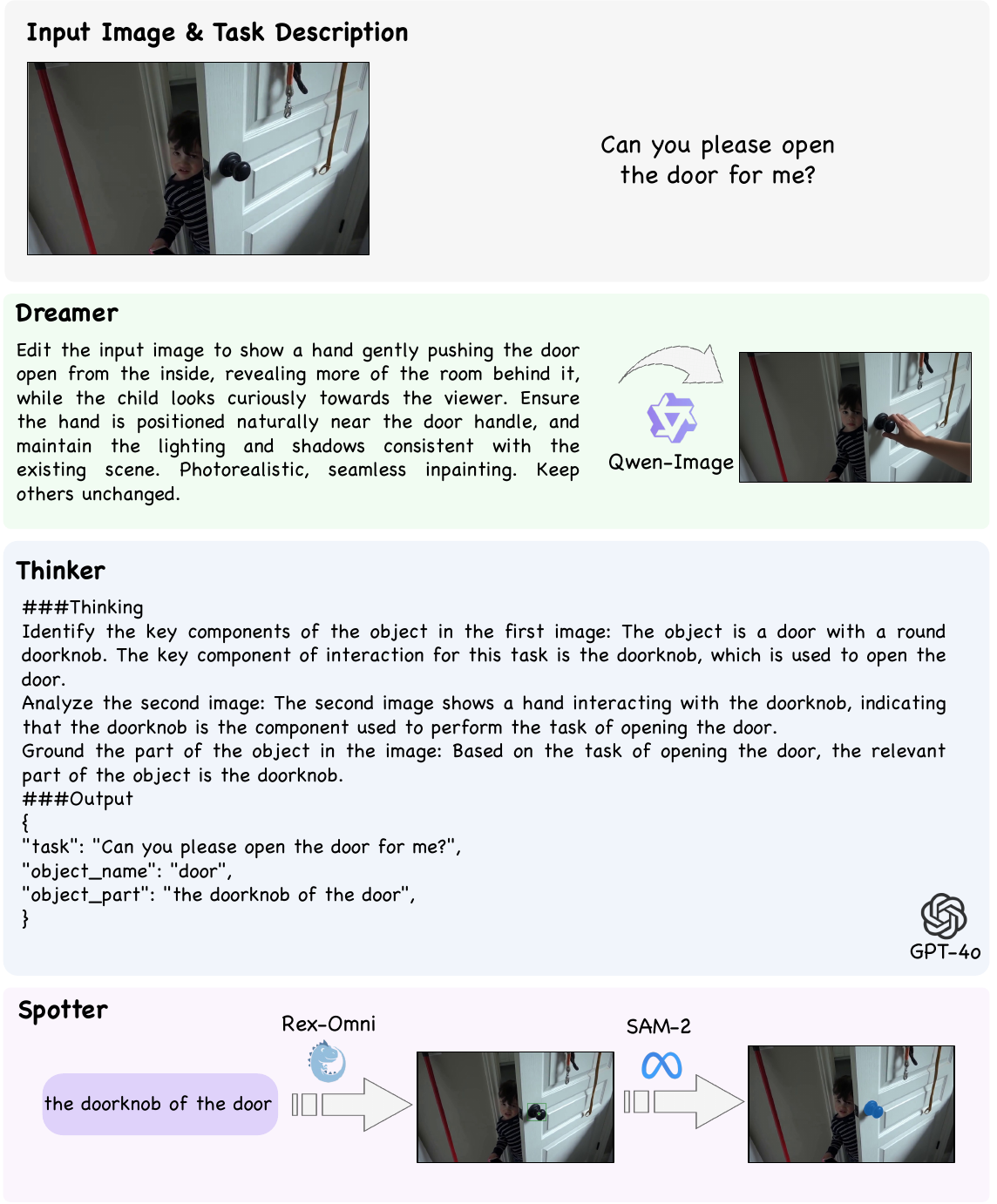}
    \caption{\textbf{Full Demonstration of Intermediate Results.} Sampled from results on the RAGNet dataset.}
    \label{fig:case5}
\end{figure*}

\begin{figure*}
    \centering
    \includegraphics[width=1\linewidth]{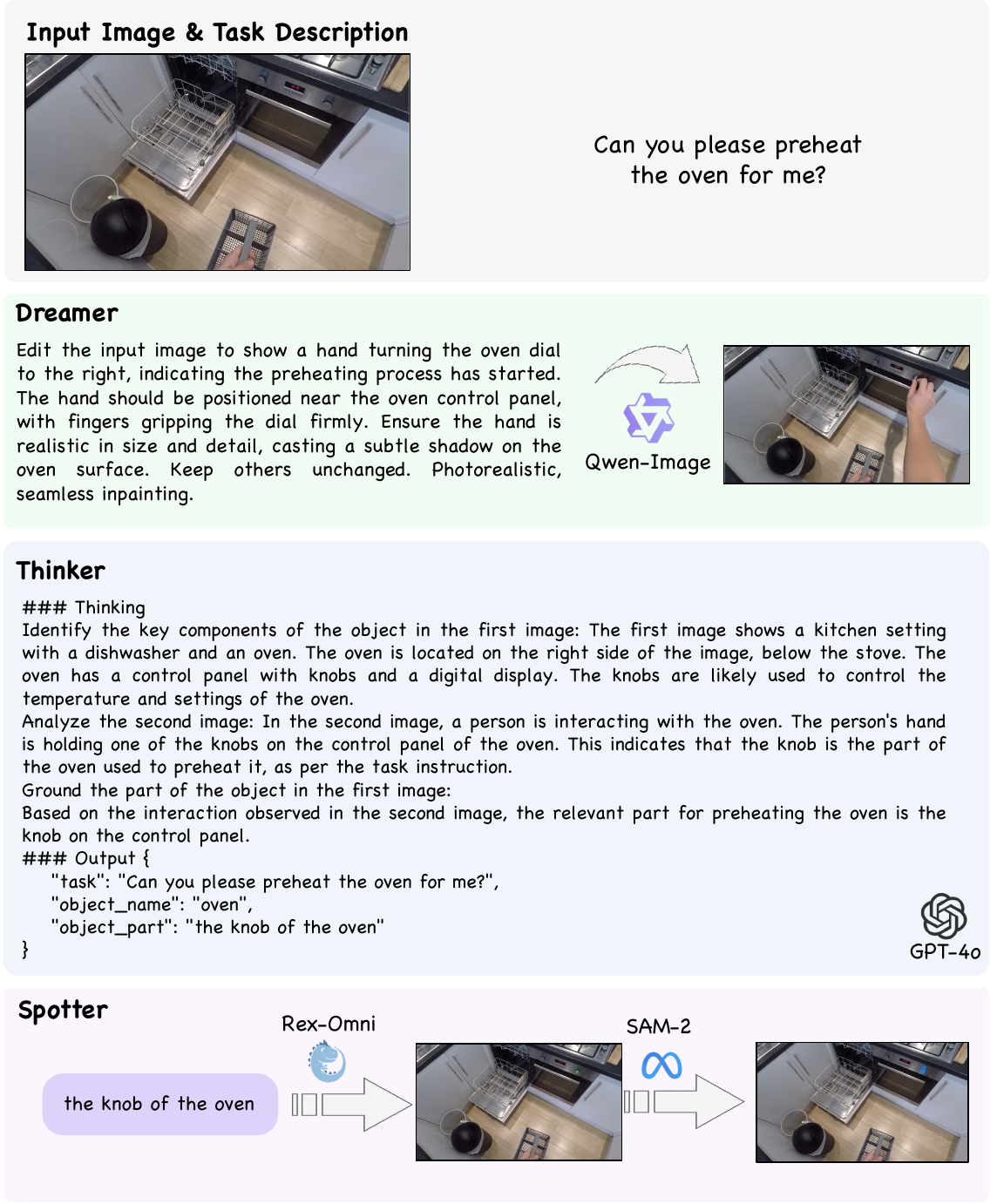}
    \caption{\textbf{Full Demonstration of Intermediate Results.} Sampled from results on the RAGNet dataset.}
    \label{fig:case6}
\end{figure*}

We here show more intermediate results of our \name~ in Figure ~\ref{fig:case1} to~\ref{fig:case6}, where Fig.~\ref{fig:case1} and Fig.~\ref{fig:case2} are sampled results on ReasonAff dataset~\cite{affordance-r1}, Fig.~\ref{fig:case3} and Fig.~\ref{fig:case4} are sampled results on UMD dataset~\cite{umd}, and Fig.~\ref{fig:case5} and Fig.~\ref{fig:case6} are sampled results on RAGNet Dataset~\cite{wu2025ragnet}.

%% file: main.bib
@inproceedings{yang2023grounding,
  title={Grounding 3d object affordance from 2d interactions in images},
  author={Yang, Yuhang and Zhai, Wei and Luo, Hongchen and Cao, Yang and Luo, Jiebo and Zha, Zheng-Jun},
  booktitle={Proceedings of the IEEE/CVF International Conference on Computer Vision},
  pages={10905--10915},
  year={2023}
}

@article{luo2023learning,
  title={Learning visual affordance grounding from demonstration videos},
  author={Luo, Hongchen and Zhai, Wei and Zhang, Jing and Cao, Yang and Tao, Dacheng},
  journal={IEEE Transactions on Neural Networks and Learning Systems},
  year={2023},
  publisher={IEEE}
}

@inproceedings{deng20213d,
  title={3d affordancenet: A benchmark for visual object affordance understanding},
  author={Deng, Shengheng and Xu, Xun and Wu, Chaozheng and Chen, Ke and Jia, Kui},
  booktitle={proceedings of the IEEE/CVF conference on computer vision and pattern recognition},
  pages={1778--1787},
  year={2021}
}

@article{ma2025glover++,
  title={GLOVER++: Unleashing the Potential of Affordance Learning from Human Behaviors for Robotic Manipulation},
  author={Ma, Teli and Zheng, Jia and Wang, Zifan and Gao, Ziyao and Zhou, Jiaming and Liang, Junwei},
  journal={arXiv preprint arXiv:2505.11865},
  year={2025}
}

@article{chu20253d,
  title={3D-AffordanceLLM: Harnessing Large Language Models for Open-Vocabulary Affordance Detection in 3D Worlds},
  author={Chu, Hengshuo and Deng, Xiang and Chen, Xiaoyang and Li, Yinchuan and Hao, Jianye and Nie, Liqiang},
  journal={arXiv preprint arXiv:2502.20041},
  year={2025}
}

@inproceedings{qian2024affordancellm,
  title={Affordancellm: Grounding affordance from vision language models},
  author={Qian, Shengyi and Chen, Weifeng and Bai, Min and Zhou, Xiong and Tu, Zhuowen and Li, Li Erran},
  booktitle={Proceedings of the IEEE/CVF Conference on Computer Vision and Pattern Recognition},
  pages={7587--7597},
  year={2024}
}

@inproceedings{yu2025seqafford,
  title={Seqafford: Sequential 3d affordance reasoning via multimodal large language model},
  author={Yu, Chunlin and Wang, Hanqing and Shi, Ye and Luo, Haoyang and Yang, Sibei and Yu, Jingyi and Wang, Jingya},
  booktitle={Proceedings of the Computer Vision and Pattern Recognition Conference},
  pages={1691--1701},
  year={2025}
}

@article{bai2025qwen2,
  title={Qwen2.5-vl technical report},
  author={Bai, Shuai and Chen, Keqin and Liu, Xuejing and Wang, Jialin and Ge, Wenbin and Song, Sibo and Dang, Kai and Wang, Peng and Wang, Shijie and Tang, Jun and others},
  journal={arXiv preprint arXiv:2502.13923},
  year={2025}
}

@inproceedings{llava,
  author={Haotian Liu and Chunyuan Li and Qingyang Wu and Yong Jae Lee},
  title={Visual Instruction Tuning},
  booktitle={Advances in Neural Information Processing Systems 2023, NeurIPS 2023},
  year={2023}
}

@article{gibson1977theory,
  title={The theory of affordances},
  author={Gibson, James J},
  journal={Hilldale, USA},
  volume={1},
  number={2},
  pages={67--82},
  year={1977}
}

@article{gao2024learning,
  title={Learning 2d invariant affordance knowledge for 3d affordance grounding},
  author={Gao, Xianqiang and Zhang, Pingrui and Qu, Delin and Wang, Dong and Wang, Zhigang and Ding, Yan and Zhao, Bin and Li, Xuelong},
  journal={arXiv preprint arXiv:2408.13024},
  year={2024}
}

@article{shao2024great,
  title={GREAT: Geometry-Intention Collaborative Inference for Open-Vocabulary 3D Object Affordance Grounding},
  author={Shao, Yawen and Zhai, Wei and Yang, Yuhang and Luo, Hongchen and Cao, Yang and Zha, Zheng-Jun},
  journal={arXiv preprint arXiv:2411.19626},
  year={2024}
}

@inproceedings{nguyen2023open,
  title={Open-vocabulary affordance detection in 3d point clouds},
  author={Nguyen, Toan and Vu, Minh Nhat and Vuong, An and Nguyen, Dzung and Vo, Thieu and Le, Ngan and Nguyen, Anh},
  booktitle={2023 IEEE/RSJ International Conference on Intelligent Robots and Systems (IROS)},
  pages={5692--5698},
  year={2023},
  organization={IEEE}
}

@misc{3DAffordSplat,
  title={3DAffordSplat: Efficient Affordance Reasoning with 3D Gaussians},
  author={Zeming wei and Junyi Lin and Yang Liu and Weixing Chen and Jingzhou Luo and Guanbin Li and Liang Lin},
  year={2025},
  eprint={2504.11218},
  archivePrefix={arXiv},
  primaryClass={cs.CV},
  url={https://arxiv.org/abs/2504.11218}
}

@article{liu2023grounding,
  title={Grounding dino: Marrying dino with grounded pre-training for open-set object detection},
  author={Liu, Shilong and Zeng, Zhaoyang and Ren, Tianhe and Li, Feng and Zhang, Hao and Yang, Jie and Jiang, Qing and Li, Chunyuan and Yang, Jianwei and Su, Hang and others},
  journal={arXiv preprint arXiv:2303.05499},
  year={2023}
}

@inproceedings{cot,
  author={Wei, Jason and Wang, Xuezhi and Schuurmans, Dale and Bosma, Maarten and ichter, brian and Xia, Fei and Chi, Ed and Le, Quoc V and Zhou, Denny},
  booktitle={Advances in Neural Information Processing Systems},
  editor={S. Koyejo and S. Mohamed and A. Agarwal and D. Belgrave and K. Cho and A. Oh},
  pages={24824--24837},
  publisher={Curran Associates, Inc.},
  title={Chain-of-Thought Prompting Elicits Reasoning in Large Language Models},
  volume={35},
  year={2022}
}

@inproceedings{wu2024minds,
  title={Mind's Eye of {LLM}s: Visualization-of-Thought Elicits Spatial Reasoning in Large Language Models},
  author={Wenshan Wu and Shaoguang Mao and Yadong Zhang and Yan Xia and Li Dong and Lei Cui and Furu Wei},
  booktitle={The Thirty-eighth Annual Conference on Neural Information Processing Systems},
  year={2024}
}

@article{generalworldmodelsurvey,
  title={Is Sora a World Simulator? A Comprehensive Survey on General World Models and Beyond},
  author={Zheng Zhu and Xiaofeng Wang and Wangbo Zhao and Chen Min and Nianchen Deng and Min Dou and Yuqi Wang and Botian Shi and Kai Wang and Chi Zhang and Yang You and Zhaoxiang Zhang and Dawei Zhao and Liang Xiao and Jian Zhao and Jiwen Lu and Guan Huang},
  journal={arXiv preprint arXiv:2405.03520},
  year={2024}
}

@article{affordance-r1,
  title={Affordance-R1: Reinforcement Learning for Generalizable Affordance Reasoning in Multimodal Large Language Model},
  author={Wang, Hanqing and Wang, Shaoyang and Zhong, Yiming and Yang, Zemin and Wang, Jiamin and Cui, Zhiqing and Yuan, Jiahao and Han, Yifan and Liu, Mingyu and Ma, Yuexin},
  journal={arXiv preprint arXiv:2508.06206},
  year={2025}
}

@article{seg-zero,
  title={Seg-zero: Reasoning-chain guided segmentation via cognitive reinforcement},
  author={Liu, Yuqi and Peng, Bohao and Zhong, Zhisheng and Yue, Zihao and Lu, Fanbin and Yu, Bei and Jia, Jiaya},
  journal={arXiv preprint arXiv:2503.06520},
  year={2025}
}

@article{visionreasoner,
  title        = {VisionReasoner: Unified Visual Perception and Reasoning via Reinforcement Learning},
  author       = {Liu, Yuqi and Qu, Tianyuan and Zhong, Zhisheng and Peng, Bohao and Liu, Shu and Yu, Bei and Jia, Jiaya},
  journal      = {arXiv preprint arXiv:2505.12081},
  year         = {2025}
}

@article{zhang2025phystoolbench,
  title={PhysToolBench: Benchmarking Physical Tool Understanding for MLLMs},
  author={Zhang, Zixin and Chen, Kanghao and Lin, Xingwang and Jiang, Lutao and Zheng, Xu and Lyu, Yuanhuiyi and Guo, Litao and Li, Yinchuan and Chen, Ying-Cong},
  journal={arXiv preprint arXiv:2510.09507},
  year={2025}
}

@misc{qwen-image,
      title={Qwen-Image Technical Report}, 
      author={Chenfei Wu and Jiahao Li and Jingren Zhou and Junyang Lin and Kaiyuan Gao and Kun Yan and Sheng-ming Yin and Shuai Bai and Xiao Xu and Yilei Chen and Yuxiang Chen and Zecheng Tang and Zekai Zhang and Zhengyi Wang and An Yang and Bowen Yu and Chen Cheng and Dayiheng Liu and Deqing Li and Hang Zhang and Hao Meng and Hu Wei and Jingyuan Ni and Kai Chen and Kuan Cao and Liang Peng and Lin Qu and Minggang Wu and Peng Wang and Shuting Yu and Tingkun Wen and Wensen Feng and Xiaoxiao Xu and Yi Wang and Yichang Zhang and Yongqiang Zhu and Yujia Wu and Yuxuan Cai and Zenan Liu},
      year={2025},
      eprint={2508.02324},
      archivePrefix={arXiv},
      primaryClass={cs.CV},
      url={https://arxiv.org/abs/2508.02324}, 
}

@misc{rex-omni,
      title={Detect Anything via Next Point Prediction}, 
      author={Qing Jiang and Junan Huo and Xingyu Chen and Yuda Xiong and Zhaoyang Zeng and Yihao Chen and Tianhe Ren and Junzhi Yu and Lei Zhang},
      year={2025},
      eprint={2510.12798},
      archivePrefix={arXiv},
      primaryClass={cs.CV},
      url={https://arxiv.org/abs/2510.12798}, 
}

@inproceedings{
  wan2024instructpart,
  title={InstructPart: Task-Oriented Part Segmentation with Instruction Reasoning},
  author={Wan, Zifu and  Xie, Yaqi and Zhang, Ce and Lin, Zhiqiu and Wang, Zihan and Stepputtis, Simon and Ramanan, Deva and Sycara, Katia},
  booktitle={The 63rd Annual Meeting of the Association for Computational Linguistics},
  year={2025},
  url={https://openreview.net/forum?id=IMEr4XgJSZ}
}

@inproceedings{umd,
  title={Affordance detection of tool parts from geometric features},
  author={Myers, Austin and Teo, Ching L and Ferm{\"u}ller, Cornelia and Aloimonos, Yiannis},
  booktitle={2015 IEEE international conference on robotics and automation (ICRA)},
  pages={1374--1381},
  year={2015},
  organization={IEEE}
}

@inproceedings{wu2025ragnet,
  title={RAGNet: Large-scale Reasoning-based Affordance Segmentation Benchmark towards General Grasping},
  author={Wu, Dongming and Fu, Yanping and Huang, Saike and Liu, Yingfei and Jia, Fan and Liu, Nian and Dai, Feng and Wang, Tiancai and Anwer, Rao Muhammad and Khan, Fahad Shahbaz and others},
  booktitle={Proceedings of the IEEE/CVF International Conference on Computer Vision},
  pages={11980--11990},
  year={2025}
}

@inproceedings{lai2024lisa,
  title={Lisa: Reasoning segmentation via large language model},
  author={Lai, Xin and Tian, Zhuotao and Chen, Yukang and Li, Yanwei and Yuan, Yuhui and Liu, Shu and Jia, Jiaya},
  booktitle={Proceedings of the IEEE/CVF Conference on Computer Vision and Pattern Recognition},
  pages={9579--9589},
  year={2024}
}

@inproceedings{chao2015hico,
  title={Hico: A benchmark for recognizing human-object interactions in images},
  author={Chao, Yu-Wei and Wang, Zhan and He, Yugeng and Wang, Jiaxuan and Deng, Jia},
  booktitle={Proceedings of the IEEE international conference on computer vision},
  pages={1017--1025},
  year={2015}
}

@inproceedings{gkioxari2018detecting,
  title={Detecting and recognizing human-object interactions},
  author={Gkioxari, Georgia and Girshick, Ross and Doll{\'a}r, Piotr and He, Kaiming},
  booktitle={Proceedings of the IEEE conference on computer vision and pattern recognition},
  pages={8359--8367},
  year={2018}
}

@article{kumar2018visual,
  title={Visual memory for robust path following},
  author={Kumar, Ashish and Gupta, Saurabh and Fouhey, David and Levine, Sergey and Malik, Jitendra},
  journal={Advances in neural information processing systems},
  volume={31},
  year={2018}
}

@article{bahl2022human,
  title={Human-to-robot imitation in the wild},
  author={Bahl, Shikhar and Gupta, Abhinav and Pathak, Deepak},
  journal={arXiv preprint arXiv:2207.09450},
  year={2022}
}

@article{hsu2023ditto,
  title={Ditto in the house: Building articulation models of indoor scenes through interactive perception},
  author={Hsu, Cheng-Chun and Jiang, Zhenyu and Zhu, Yuke},
  journal={arXiv preprint arXiv:2302.01295},
  year={2023}
}

@article{yang2025qwen3,
  title={Qwen3 technical report},
  author={Yang, An and Li, Anfeng and Yang, Baosong and Zhang, Beichen and Hui, Binyuan and Zheng, Bo and Yu, Bowen and Gao, Chang and Huang, Chengen and Lv, Chenxu and others},
  journal={arXiv preprint arXiv:2505.09388},
  year={2025}
}

@article{gpt4,
  title={Gpt-4 technical report},
  author={Achiam, Josh and Adler, Steven and Agarwal, Sandhini and Ahmad, Lama and Akkaya, Ilge and Aleman, Florencia Leoni and Almeida, Diogo and Altenschmidt, Janko and Altman, Sam and Anadkat, Shyamal and others},
  journal={arXiv preprint arXiv:2303.08774},
  year={2023}
}

@misc{openaio1,
  author    = {OpenAI},
  title     = {{OpenAI o1}},
  howpublished = {\url{https://openai.com/o1/}},
  year      = {2024}
}

@article{guo2025deepseekr1,
  title={Deepseek-r1: Incentivizing reasoning capability in llms via reinforcement learning},
  author={Guo, Daya and Yang, Dejian and Zhang, Haowei and Song, Junxiao and Zhang, Ruoyu and Xu, Runxin and Zhu, Qihao and Ma, Shirong and Wang, Peiyi and Bi, Xiao and others},
  journal={arXiv preprint arXiv:2501.12948},
  year={2025}
}

@article{shao2024deepseekmath,
  title={Deepseekmath: Pushing the limits of mathematical reasoning in open language models},
  author={Shao, Zhihong and Wang, Peiyi and Zhu, Qihao and Xu, Runxin and Song, Junxiao and Bi, Xiao and Zhang, Haowei and Zhang, Mingchuan and Li, YK and Wu, Y and others},
  journal={arXiv preprint arXiv:2402.03300},
  year={2024}
}

@article{Shen2025VLMR1AS,
  title={VLM-R1: A Stable and Generalizable R1-style Large Vision-Language Model},
  author={Haozhan Shen and Peng Liu and Jingcheng Li and Chunxin Fang and Yibo Ma and Jiajia Liao and Qiaoli Shen and Zilun Zhang and Kangjia Zhao and Qianqian Zhang and Ruochen Xu and Tiancheng Zhao},
  journal={ArXiv},
  year={2025},
  volume={abs/2504.07615},
  url={https://api.semanticscholar.org/CorpusID:277667819}
}

@article{Huang2025VisionR1IR,
  title={Vision-R1: Incentivizing Reasoning Capability in Multimodal Large Language Models},
  author={Wenxuan Huang and Bohan Jia and Zijie Zhai and Shaoshen Cao and Zheyu Ye and Fei Zhao and Zhe Xu and Yao Hu and Shaohui Lin},
  journal={ArXiv},
  year={2025},
  volume={abs/2503.06749},
  url={https://api.semanticscholar.org/CorpusID:276902576}
}

@article{ravi2024sam,
  title={Sam 2: Segment anything in images and videos},
  author={Ravi, Nikhila and Gabeur, Valentin and Hu, Yuan-Ting and Hu, Ronghang and Ryali, Chaitanya and Ma, Tengyu and Khedr, Haitham and R{\"a}dle, Roman and Rolland, Chloe and Gustafson, Laura and others},
  journal={arXiv preprint arXiv:2408.00714},
  year={2024}
}

@article{li2025imagine,
  title={Imagine while reasoning in space: Multimodal visualization-of-thought},
  author={Li, Chengzu and Wu, Wenshan and Zhang, Huanyu and Xia, Yan and Mao, Shaoguang and Dong, Li and Vuli{\'c}, Ivan and Wei, Furu},
  journal={arXiv preprint arXiv:2501.07542},
  year={2025}
}

@inproceedings{li2023locate,
  title={Locate: Localize and transfer object parts for weakly supervised affordance grounding},
  author={Li, Gen and Jampani, Varun and Sun, Deqing and Sevilla-Lara, Laura},
  booktitle={Proceedings of the IEEE/CVF Conference on Computer Vision and Pattern Recognition},
  pages={10922--10931},
  year={2023}
}

@inproceedings{luo2022learning,
  title={Learning affordance grounding from exocentric images},
  author={Luo, Hongchen and Zhai, Wei and Zhang, Jing and Cao, Yang and Tao, Dacheng},
  booktitle={Proceedings of the IEEE/CVF conference on computer vision and pattern recognition},
  pages={2252--2261},
  year={2022}
}

@article{hurst2024gpt,
  title={Gpt-4o system card},
  author={Hurst, Aaron and Lerer, Adam and Goucher, Adam P and Perelman, Adam and Ramesh, Aditya and Clark, Aidan and Ostrow, AJ and Welihinda, Akila and Hayes, Alan and Radford, Alec and others},
  journal={arXiv preprint arXiv:2410.21276},
  year={2024}
}

@inproceedings{vlpart,
  title={Going denser with open-vocabulary part segmentation},
  author={Sun, Peize and Chen, Shoufa and Zhu, Chenchen and Xiao, Fanyi and Luo, Ping and Xie, Saining and Yan, Zhicheng},
  booktitle={Proceedings of the IEEE/CVF International Conference on Computer Vision},
  pages={15453--15465},
  year={2023}
}

@inproceedings{ovseg,
  title={Open-vocabulary semantic segmentation with mask-adapted clip},
  author={Liang, Feng and Wu, Bichen and Dai, Xiaoliang and Li, Kunpeng and Zhao, Yinan and Zhang, Hang and Zhang, Peizhao and Vajda, Peter and Marculescu, Diana},
  booktitle={Proceedings of the IEEE/CVF conference on computer vision and pattern recognition},
  pages={7061--7070},
  year={2023}
}

@inproceedings{SAN,
  title={Side adapter network for open-vocabulary semantic segmentation},
  author={Xu, Mengde and Zhang, Zheng and Wei, Fangyun and Hu, Han and Bai, Xiang},
  booktitle={Proceedings of the IEEE/CVF conference on computer vision and pattern recognition},
  pages={2945--2954},
  year={2023}
}

@inproceedings{chen2024sam4mllm,
  title={SAM4MLLM: Enhance Multi-Modal Large Language Model for Referring Expression Segmentation},
  author={Chen, Yi-Chia and Li, Wei-Hua and Sun, Cheng and Wang, Yu-Chiang Frank and Chen, Chu-Song},
  booktitle={European Conference on Computer Vision},
  pages={323--340},
  year={2024},
  organization={Springer}
}

@misc{internvl3,
  title        = {InternVL3: Exploring Advanced Training and Test-Time Recipes for Open-Source Multimodal Models},
  author       = {Jinguo Zhu and Weiyun Wang and Zhe Chen and Zhaoyang Liu and Shenglong Ye and Lixin Gu and Hao Tian and Yuchen Duan and others},
  year         = {2025},
  eprint       = {2504.10479},
  archivePrefix= {arXiv},
  primaryClass = {cs.CV},
  url          = {https://arxiv.org/abs/2504.10479}
}

@inproceedings{glamm,
  title={Glamm: Pixel grounding large multimodal model},
  author={Rasheed, Hanoona and Maaz, Muhammad and Shaji, Sahal and Shaker, Abdelrahman and Khan, Salman and Cholakkal, Hisham and Anwer, Rao M and Xing, Eric and Yang, Ming-Hsuan and Khan, Fahad S},
  booktitle={Proceedings of the IEEE/CVF Conference on Computer Vision and Pattern Recognition},
  pages={13009--13018},
  year={2024}
}

@article{chen2025tivibench,
  title={TiViBench: Benchmarking Think-in-Video Reasoning for Video Generative Models},
  author={Chen, Harold Haodong and Lan, Disen and Shu, Wen-Jie and Liu, Qingyang and Wang, Zihan and Chen, Sirui and Cheng, Wenkai and Chen, Kanghao and Zhang, Hongfei and Zhang, Zixin and others},
  journal={arXiv preprint arXiv:2511.13704},
  year={2025}
}

@misc{chernThinkingGeneratedImages2025,
  title = {Thinking with {{Generated Images}}},
  author = {Chern, Ethan and Hu, Zhulin and Chern, Steffi and Kou, Siqi and Su, Jiadi and Ma, Yan and Deng, Zhijie and Liu, Pengfei},
  year = 2025,
  number = {arXiv:2505.22525},
  publisher = {arXiv},
  doi = {10.48550/arXiv.2505.22525},
  archiveprefix = {arXiv}
}

@article{zhang2025dualcamctrl,
  title={DualCamCtrl: Dual-Branch Diffusion Model for Geometry-Aware Camera-Controlled Video Generation},
  author={Zhang, Hongfei and Chen, Kanghao and Zhang, Zixin and Chen, Harold Haodong and Lyu, Yuanhuiyi and Zhang, Yuqi and Yang, Shuai and Zhou, Kun and Chen, Yingcong},
  journal={arXiv preprint arXiv:2511.23127},
  year={2025}
}

@article{wan2025wan,
  title={Wan: Open and advanced large-scale video generative models},
  author={Wan, Team and Wang, Ang and Ai, Baole and Wen, Bin and Mao, Chaojie and Xie, Chen-Wei and Chen, Di and Yu, Feiwu and Zhao, Haiming and Yang, Jianxiao and others},
  journal={arXiv preprint arXiv:2503.20314},
  year={2025}
}

@article{guo2025video,
  title={Are Video Models Ready as Zero-Shot Reasoners? An Empirical Study with the MME-CoF Benchmark},
  author={Guo, Ziyu and Chen, Xinyan and Zhang, Renrui and An, Ruichuan and Qi, Yu and Jiang, Dongzhi and Li, Xiangtai and Zhang, Manyuan and Li, Hongsheng and Heng, Pheng-Ann},
  journal={arXiv preprint arXiv:2510.26802},
  year={2025}
}

@article{guo2025comfymind,
  title={ComfyMind: Toward General-Purpose Generation via Tree-Based Planning and Reactive Feedback},
  author={Guo, Litao and Xu, Xinli and Wang, Luozhou and Lin, Jiantao and Zhou, Jinsong and Zhang, Zixin and Su, Bolan and Chen, Ying-Cong},
  journal={arXiv preprint arXiv:2505.17908},
  year={2025}
}

@inproceedings{chen2025hierarchical,
title={Hierarchical Fine-grained Preference Optimization for Physically Plausible Video Generation},
author={Harold Haodong Chen and Haojian Huang and Qifeng Chen and Harry Yang and Ser-Nam Lim},
booktitle={The Thirty-ninth Annual Conference on Neural Information Processing Systems},
year={2025},
url={https://openreview.net/forum?id=y0SRR9XGlZ}
}

@inproceedings{zhong2025omnisam,
  title={Omnisam: Omnidirectional segment anything model for uda in panoramic semantic segmentation},
  author={Zhong, Ding and Zheng, Xu and Liao, Chenfei and Lyu, Yuanhuiyi and Chen, Jialei and Wu, Shengyang and Zhang, Linfeng and Hu, Xuming},
  booktitle={Proceedings of the IEEE/CVF International Conference on Computer Vision},
  pages={23892--23901},
  year={2025}
}
